\documentclass[10pt,twocolumn,letterpaper]{article}

\usepackage[pagenumbers]{wacv} 

\usepackage{graphicx}
\usepackage{amsmath}
\usepackage{amssymb}
\usepackage{booktabs}
\usepackage{multirow}
\usepackage[algo2e]{algorithm2e} 
\usepackage{algorithm}
\usepackage{algpseudocode}
\usepackage{diagbox}

\usepackage[pagebackref,breaklinks,colorlinks]{hyperref}

\usepackage[capitalize]{cleveref}
\crefname{section}{Sec.}{Secs.}
\Crefname{section}{Section}{Sections}
\Crefname{table}{Table}{Tables}
\crefname{table}{Tab.}{Tabs.}

\def\wacvPaperID{876} 
\def\confName{WACV}
\def\confYear{2025}

\begin{document}

\title{Shadow Removal Refinement via Material-Consistent Shadow Edges}

\author{Shilin Hu\textsuperscript{1}, Hieu Le\textsuperscript{2}, ShahRukh Athar\textsuperscript{1}, Sagnik Das\textsuperscript{3}, Dimitris Samaras\textsuperscript{1}\\
\textsuperscript{1}Stony Brook University \hspace{10pt} \textsuperscript{2}EPFL \hspace{10pt} \textsuperscript{3}Amazon \\
}

\maketitle

\begin{abstract}
Shadow boundaries can be confused with material boundaries as both exhibit sharp changes in luminance or contrast within a scene. However, shadows do not modify the intrinsic color or texture of surfaces.  
Therefore, on both sides of shadow edges traversing regions with the same material, the original color and textures should be the same if the shadow is removed properly. 
These shadow/shadow-free pairs are very useful but hard-to-collect supervision signals.
The crucial contribution of this paper is to learn how to identify those shadow edges that traverse material-consistent regions and how to use them as self-supervision for shadow removal refinement during test time. To achieve this, we fine-tune SAM, an image segmentation foundation model, to produce a shadow-invariant segmentation and then extract material-consistent shadow edges by comparing the SAM segmentation with the shadow mask. 
Utilizing these shadow edges, we introduce color and texture-consistency losses to enhance the shadow removal process. We demonstrate the effectiveness of our method in improving shadow removal results on more challenging, in-the-wild images, outperforming the state-of-the-art shadow removal methods. Additionally, we propose a new metric and an annotated dataset for evaluating the performance of shadow removal methods without the need for paired shadow/shadow-free data.
\end{abstract}

\section{Introduction}
\label{sec:intro}

\begin{figure*}[!t]
    \centering
    \includegraphics[width=\linewidth]{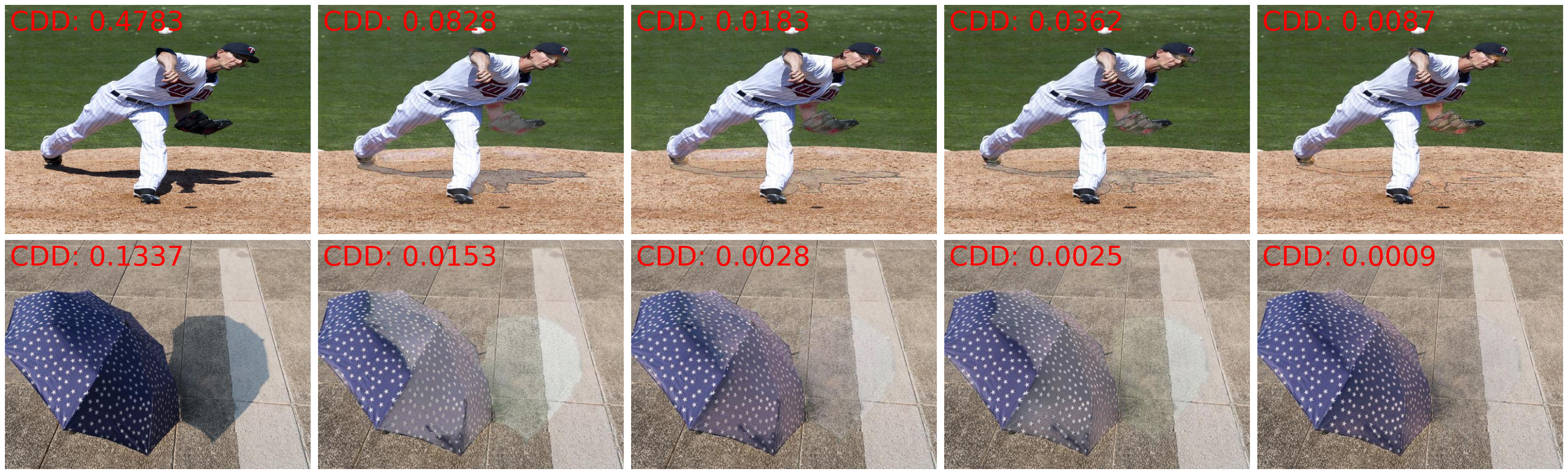}
    \makebox[0.19\linewidth]{\small{(a) Input}}
    \makebox[0.19\linewidth]{\small{(b) SID \cite{Le_2019_ICCV}}}
    \makebox[0.19\linewidth]{\small{(c) SID+Ours}}
    \makebox[0.19\linewidth]{\small{(d) ShadowFormer\cite{guo2023shadowformer}}}
    \makebox[0.19\linewidth]{\small{(e) ShadowFormer+Ours}}
    \caption{Examples from our proposed SBU-S (\textit{top}) and CUHK-S (\textit{bottom}) testing sets. We show the shadow removal results of two existing state-of-the-art methods, SID\cite{Le_2019_ICCV} and ShadowFormer\cite{guo2023shadowformer} in columns (b) and (d) respectively for two challenging cases. The results of both methods are significantly improved when used jointly with our refinement method - as in columns (c) and (e). Our proposed Color Distribution Difference (CDD) metric for each image is shown in red, which can measure shadow removal performance without the need for shadow-free images.}
    \label{fig:teaser}
    \vspace{-3mm}
\end{figure*}

Shadow edges were extensively studied in early shadow removal work\cite{Liu2008TextureConsistentSR, finlayson2005removal, xu2006shadow}. These edges delineate the boundary between shadow and non-shadow areas, providing a clear indication of the transition in the pixel color and intensity that characterizes the shadow, providing useful supervision signals for shadow removal \cite{wu2007natural,khanPami15,Le_2020_ECCV}.
However, shadow edges can also coincide with object boundaries. In these cases, the color and texture differences between the two sides of the shadow edges result from both the shadow effect and the different materials, which are hard to disentangle. Therefore, identifying shadow edges that cross areas of consistent material is beneficial for shadow removal. However, this is challenging since object boundaries can be confused with shadow boundaries as both exhibit sharp changes in luminance or contrast within a scene. In this work, we propose a novel method to identify those material-consistent shadow edges and use them as supervision signals for shadow removal. 


To identify material-consistent shadow edges, we propose to fine-tune SAM \cite{kirillov2023segany}, an image segmentation foundation model, to produce a shadow-invariant segmentation. The main idea is to force SAM to output the same segmentation mask, with or without shadows, thereby guiding the model to be less responsive to the shadow presence. 
We then compare the output shadow-invariant segmentation with the shadow mask to obtain material-consistent shadow edges. These are shadow edges that do not coincide with any material edges from SAM. 

These material-consistent shadow edges can be extracted for any image. We show that at test time, we can extract these edges and use them as supervision to refine shadow-removal results. To do so, we sample pixels and patches alongside the shadow edges to form shadow/shadow-free pairs. We then introduce two losses: a RGB distance loss and a RGB distribution loss. The RGB distance loss calculates the minimum distance between each sampled shadow and non-shadow pixel along the selected edges to recover the correct color of the shadow region.  The RGB distribution loss computes the Earth Mover's Distance (EMD) between the color distributions of the sampled pixels to ensure texture consistency alongside the shadow edges. We further use the Learned Perceptual Image Patch Similarity (LPIPS) \cite{zhang2018perceptual} loss on sampled patches within the same material to constrain the texture consistency between the non-edge shadow/non-shadow region.
By imposing these constraints, we refine the pre-trained model in a \textit{self-supervised} manner, enabling adaptation to more complex shadow images (see \cref{fig:teaser}).


Further, we propose a novel evaluation metric based on material-consistent shadow edges, namely Color Distribution Difference (CDD). In essence, we can evaluate shadow removal performance by measuring the disparity in pixel color distribution on both sides of the material-consistent shadow edges. These edges can be easily annotated, even in cases where shadow-free images are hard to obtain. A lower CDD value corresponds to a more effective shadow removal performance, signifying a closer alignment between the textures on either side of the shadow edge, as shown in \cref{fig:teaser}. This new evaluation scheme enables bench-marking shadow-removal methods on challenging, in-the-wild shadow images. We curate a test set sourced from existing shadow detection datasets, namely SBU \cite{vicente2016large} and CUHK \cite{hu2021revisiting}. 
These datasets contain images under complex shadow scenarios, providing a more comprehensive representation of shadow images in general. 
We annotate pixels on both sides of the shadow edge that belong to the same background material for each image within the proposed dataset. 
This benchmark test set (to be released upon publication) will serve as a valuable resource for evaluating the generalizability of shadow removal methods on complex shadow images.


To summarize, our contributions are as follows:
\vspace{-1.0mm}
\begin{itemize}
    \item We introduce an efficient shadow edge extraction module designed to identify shadow edges that traverse the same material. We achieve this by fine-tuning SAM to generate shadow-invariant segmentation and then comparing this segmentation with the shadow mask to perform the extraction.
    \item We propose a test-time adaptation method that refines the shadow removal result, which relies on the extracted self-supervision signal to enforce the material consistency between shadow and non-shadow regions. 
    \item We curate a dataset featuring shadow images in general scenes, serving as a benchmark for assessing shadow removal methods in complex scenarios. We propose a novel evaluation metric, Color Distribution Difference (CDD) to assess the shadow removal performance, even when shadow-free ground truth is not available. 
    \item Experimental results demonstrate that our method can be seamlessly integrated with existing models, significantly enhancing performance on complex shadow images. Specifically, when combined with two state-of-the-art shadow removal approaches, SP+M-Net \cite{Le_2019_ICCV} and ShadowFormer \cite{guo2023shadowformer}, our method outperforms by at least 30\% on CDD measurement on our proposed dataset. Code will be released upon acceptance.
\end{itemize}

\section{Related Works}

Early-stage shadow removal research \cite{Finlayson09,Finlayson01,Finlayson02,Drew03recoveryof} is motivated by physical modeling of illumination and color, typically in a light source-occluder system \cite{Barrow1978,huang11,Shor08}.
The aim was to find useful self-supervision signals to fit the model and then remove the shadows. 
Guo \etal \cite{guoPami} proposed identifying pairs of regions under different illuminations within the same material.
Several methods \cite{Rubin81,Witkin82,wu2007natural,khanPami15} looked for cues at shadow edges, which typically involve designing hand-crafted features that capture the illumination and color changes. 
However, these methods are built under ideal assumptions from the physical modeling, which does not fit real-world settings due to the complexity and variability of shadow appearances. 

State-of-the-art shadow removal methods include deep networks trained end-to-end on pairs of shadow/shadow-free images \cite{wang2018stacked,le2021physics,cun2020towards,fu2021auto,chen2021canet,zhu2022bijective,guo2023shadowformer,li2023leveraging}, taking advantage of the powerful ability in learning mappings from the training pairs. 
Iterative refinement is also adopted for shadow removal, \eg ARGAN \cite{Ding2019ARGANAR}, which removed shadows progressively using a multi-step generator, each step refining the output by removing remaining shadows.
Recently, \cite{jin2022des3,mei2024latent,guo2023shadowdiffusion,liu2024recasting} have utilized diffusion models for shadow removal via iterative denoising.
A method similar to us is the work of Guo \etal \cite{guo2023boundary} that also explored boundary cues for shadow removal. They found that their illumination model does not sufficiently model the boundary region, which necessitate extra supervision during training.

A few methods try to mitigate the dependence on paired data. 
Hu \etal \cite{hu_iccv2019mask} proposed Mask-ShadowGAN, which leverages unpaired data to learn the adaptation from the shadow-free domain to the shadow domain and vice versa. 
Le and Samaras \cite{Le_2020_ECCV} introduced cropping unpaired patches from the same shadow image without the requirement of a shadow-free image. 
However, these un-/semi-supervised methods do not surpass their fully-supervised counterparts.

Our method is among the first to employ test-time adaptation (TTA) for shadow removal. In general, TTA\cite{sun2020test} aims to enhance a pre-trained model’s performance on specific test data. Xiao \etal \cite{xiao2023energybased} and Yuan \etal \cite{yuan2024tea} proposed energy-based models to align target samples with the source distribution. MEMO \cite{zhang2022memo} augmented test points in various ways to encourage consistent and confident predictions for test-time robustness. TENT \cite{wang2021tent} focused on the fully test-time adaptation setting, similar to ours, using only test data and a specific test loss for adaptation. In our case, we search for supervision from each testing sample for test-time adaption.
Leveraging the learning capacity of deep models and the guidance provided by material-consistent edges, we propose extracting self-supervision signals for refining shadow removal performance during inference.


\section{Method}
In this section, we describe our novel self-supervised test-time adaptation method for refining shadow removal, given the shadow image and its corresponding shadow mask.
First, we introduce our shadow edge extraction module to extract useful supervision from shadow/non-shadow pairs alongside shadow edges. We fine-tune the Segment Anything Model (SAM) \cite{kirillov2023segany} to learn to generate a segmentation of materials regardless of the presence of shadows. By comparing the segmentation masks with the shadow mask, we can identify edges that traverse the same materials.
Next, we outline our adaptation process, which uses the shadow edges to iteratively refine shadow removal results during inference. Comparing pixels across two sides of shadow edges provides us with information about the color transition caused by the shadow. Additionally, patches within each material mask should exhibit consistent textures when the shadow is properly removed.
Using the collected supervision from shadow/shadow-free pairs and an iterative adaptation approach, our method significantly enhances the performance of pre-trained shadow removal models, even on more complex shadow images in the wild.


\subsection{Material-Consistent Shadow Edge Extraction}
\label{sec:edgeextract}
Shadow edges mark the boundary between shadow and shadow-free areas. The areas on either side of these edges offer potential supervision for shadow removal, as they contain regions with matching colors and textures.
However, shadow edges do not always separate areas with consistent underlying materials. For instance, when an object is well-illuminated and casts a shadow on the background, shadow edges form along the object boundaries. In such cases, the two sides of the shadow edges display disparate colors and textures, lacking material consistency and thus rendering them unsuitable for supervising shadow removal.
A simple strategy to avoid selecting those edges is to segment materials within the image, and then identify shadow edges that go across the same material.

\begin{figure}[!t]
   \centering
   \includegraphics[width=\linewidth]{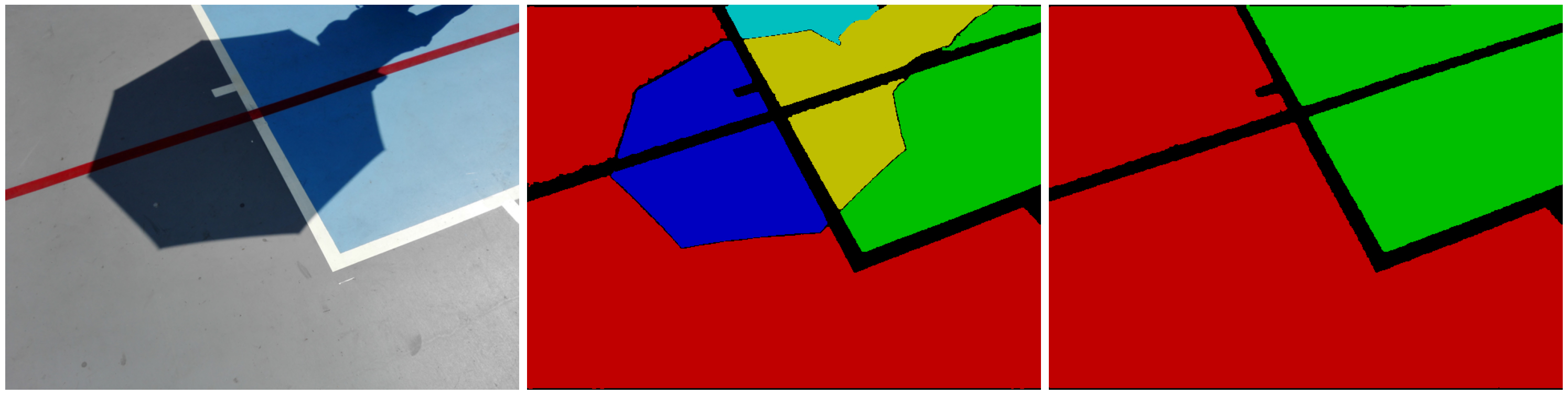}
    \makebox[0.32\linewidth]{\small{Input}}
    \makebox[0.32\linewidth]{\small{Vanilla SAM}}
    \makebox[0.32\linewidth]{\small{Fine-tuned SAM}}
   \caption{\textbf{Fine-tuning SAM for Material-Consistent Edge Extraction.} Given the input image, we compare the segmentation results of vanilla SAM \cite{kirillov2023segany} with our fine-tuned SAM. Our fine-tuned SAM achieves shadow-invariant segmentation, preserving the material consistency of each mask. In contrast, vanilla SAM is sensitive to shadow presence, segmenting shadow regions as individual masks.}
   \label{fig:sam}
   \vspace{-3mm}
\end{figure}

We propose fine-tuning the Segment Anything Model (SAM) to generate a segmentation of materials in images, irrespective of shadow presence. While Vanilla SAM demonstrates robust performance in semantic segmentation, its sensitivity to shadows results in segmenting partially shadowed materials into separate masks (\cref{fig:sam}). To mitigate this, we follow the common strategy \cite{chen2023sam,Xie_2024_WACV} of freezing the image and prompt encoders while fine-tuning the mask decoder. 
Unlike \cite{chen2023sam}, which trains adapted prompt encoders, our input prompts remain a grid of points uniformly distributed over the image (as pre-defined in \cite{kirillov2023segany}) and the input is the shadow image. 
We fine-tune SAM to be shadow-invariant by forcing it to output the same segmentation as the pseudo-ground truth generated from the shadow-free counterpart.
We use the Dice Loss \cite{sudre2017generalised} to constrain the similarity between masks predicted with and without shadows and achieve shadow-invariant material segmentation with our fine-tuned SAM, as illustrated in \cref{fig:sam}.

For a given test image, we compare the shadow mask with each of the SAM segments. The SAM is fine-tuned to ensure that each segment corresponds to a specific material. Therefore, any shadow edge intersecting a SAM segment is regarded as a material-consistent shadow edge.
This approach enables us to extract edges that traverse the same material, facilitating the sampling of pixels near the edges and patches in the shadow/non-shadow regions.

\subsection{Test-time Adaptation}
\label{sec:adaptation}
Due to the complexity and variability in shadow appearances, existing methods often struggle when confronted with out-of-distribution shadow images, leading to residual shadow effects. 
To address this challenge, we propose an iterative adaptation approach for pre-trained models based on the supervision captured from shadow edges and shadow-invariant segmentation masks.

\begin{figure}[!t]
    \centering
    \includegraphics[width=\linewidth]{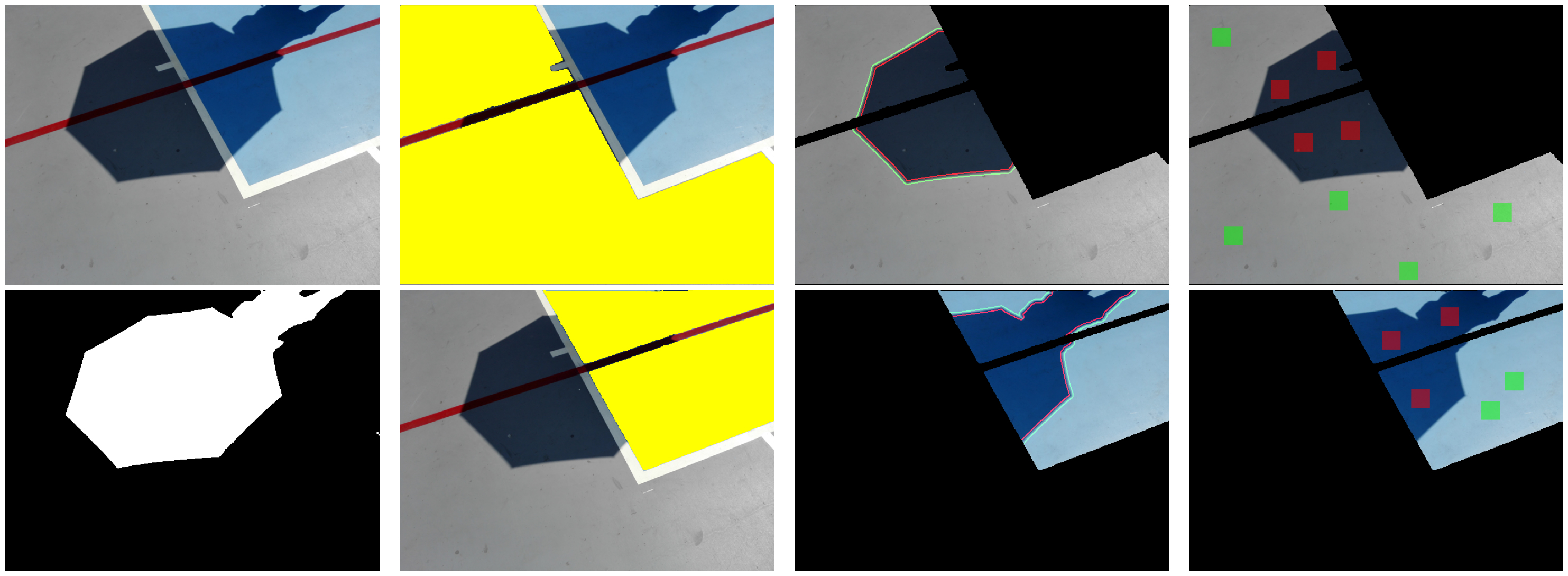}
    \makebox[0.24\linewidth]{\small{Input\&Mask}}
    \makebox[0.24\linewidth]{\small{MC Masks}}
    \makebox[0.24\linewidth]{\small{MC Pixels}}
    \makebox[0.24\linewidth]{\small{MC Patches}}
    \caption{\textbf{Supervision for the adaptation.} Given the input image and the shadow mask, we first use the fine-tuned SAM to produce material-consistent (MC) masks. We sample pixels on both sides alongside the MC shadow edge, denoted as $S_{in}$ (shown in red) and $S_{out}$ (shown in green), and patches within the same material, denoted as $P_{in}$ (shown in red) and $P_{out}$ (shown in green). Pixels and patches on both sides of the shadow edge provide supervision for the following adaptation process.}
    \label{fig:sample}
    \vspace{-3mm}
\end{figure}

To ensure color consistency between the shadow and non-shadow regions, we sample pixels along the extracted shadow edges, denoted as \(S_{in}\) and \(S_{out}\) (see \cref{fig:sample}).
$S_{in}=\{u_1,\dots,u_i,\dots,u_M\}, i\in [1, M]$ represents the pixels inside the shadow boundary, obtained by subtracting an eroded shadow mask from the original shadow mask.
Similarly, $S_{out}=\{v_1,\dots,v_j,\dots,v_N\}, j\in [1, N]$ is obtained by subtracting the shadow mask from its dilated version.
To mitigate the influence of annotation errors in the shadow mask, we apply dilation and erosion operations to the shadow mask before subtraction.
We then introduce two novel losses to constrain color consistency: 1) The RGB distance loss calculates the mean of the minimum distance between a pixel in $S_{in}$ and all pixels in $S_{out}$, enforcing the restoration of the correct color of the shadow region; 2) The RGB distribution loss computes the Earth Mover’s Distance (EMD) between the color distributions of the two sets of sampled pixels, ensuring color distribution consistency within the sampled pixel sets. The losses are formulated as:
\begin{equation}
    L_{distance} = \frac{\sum_{i=1}^M \min_{j \in [1,N]}d(u_i,v_j)}{M}, 
\end{equation}
\begin{equation}
   L_{distribution} = \textrm{EMD}([Hist(S_{in}), Hist(S_{out})]), 
\end{equation}
where $d(u,v)$ denotes the Euclidean distance between two pixels in RGB color space, and $Hist(S)$ is the histogram of sampled pixel colors across RGB channels.

To ensure texture consistency between the shadow and non-shadow regions, we randomly sample $16\times16$ patches within the segmentation mask of the same material, denoted as $P_{in}$ and $P_{out}$ (see \cref{fig:sample}).
$P_{in}=\{p_1,\dots,p_i,\dots,p_M\}, i\in [1, M]$ represents patches in the shadow region, $P_{out}=\{q_1,\dots,q_j,\dots,q_N\}, j\in [1, N]$ represents patches in the non-shadow region.
Note that, to ensure the patches are within the material masks, we apply an erosion operation to the material mask before sampling.
We then compute the mean of the minimum of the Learned Perceptual Image Patch Similarity (LPIPS) \cite{zhang2018perceptual} loss between each patch in the shadow region and all patches in the non-shadow region. The loss is formulated as follows:
\begin{equation}
    L_{per} = \frac{\sum_{i=1}^M \min_{j \in [1,N]}\textrm{LPIPS}(p_i,q_j)}{M},
\end{equation}


We compute the average across all material masks for $L_{distance}, L_{distribution}$, and $L_{per}$.
Finally, the pre-trained shadow removal model is updated iteratively to adapt to each testing case, using the weighted sum of all four losses:
\begin{equation}
    L_{total} = \lambda_1 \cdot L_{distance} + \lambda_2 \cdot L_{distribution} + \lambda_3 \cdot L_{per} 
\end{equation}

Our method can be seamlessly integrated into existing pre-trained models. The shadow edge extraction serves as a standalone module, while the proposed losses replace the training objective of the pre-trained models.
\section{New Benchmark Testset and Evaluation Metric}
The most commonly used datasets for shadow removal training and evaluation are ISTD\cite{wang2018stacked}, ISTD+\cite{Le_2019_ICCV}, and SRD \cite{qu2017deshadownet}, which provide triplets of shadow image, shadow mask, and shadow-free ground truth. However, these datasets contain only simple shadows as they lack occluders within the image. 
We argue that the current evaluation data is insufficient to assess the adaptability of current shadow removal methods due to limitations in data diversity.

\begin{figure}[!t]
  \centering
  \begin{subfigure}{0.49\linewidth}
    \includegraphics[width=\linewidth]{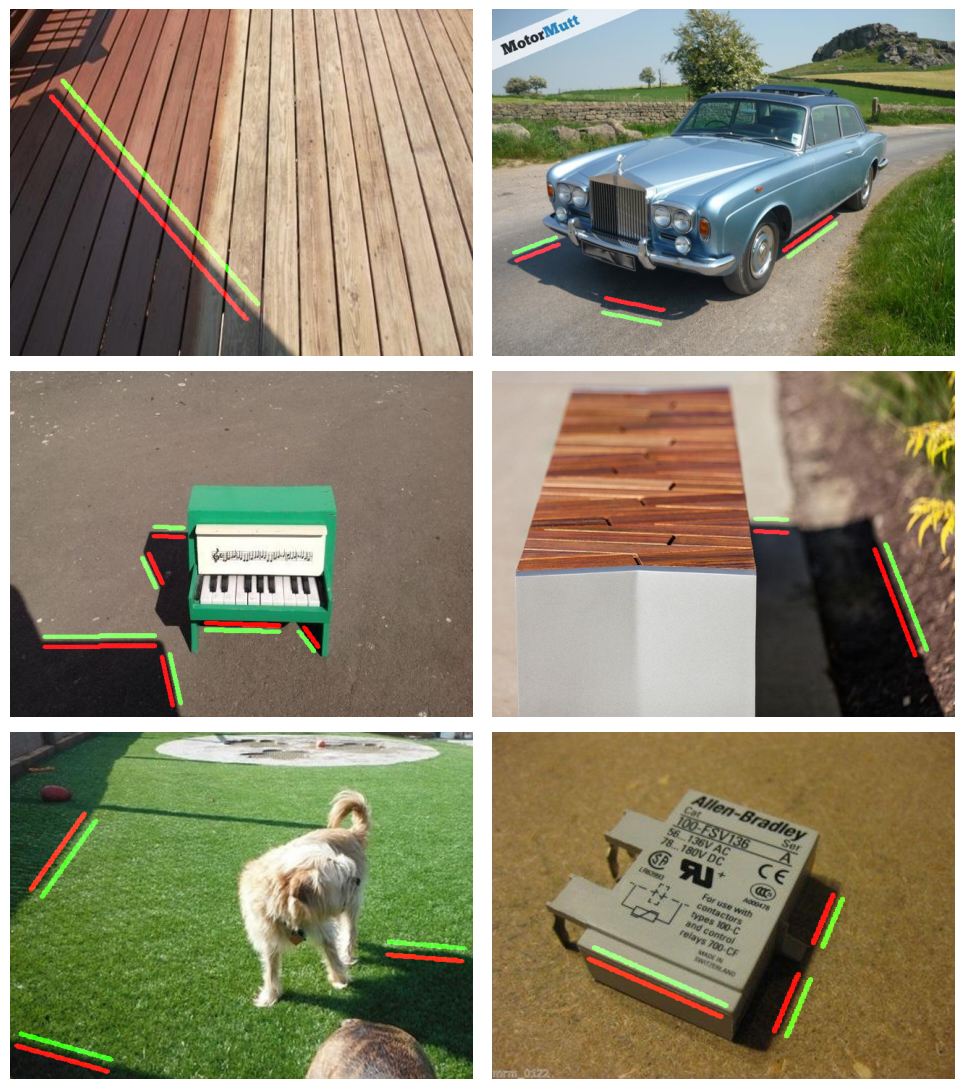}
    \caption{Shadow Images from SBU-S}
  \end{subfigure}
  \begin{subfigure}{0.49\linewidth}
    \includegraphics[width=\linewidth]{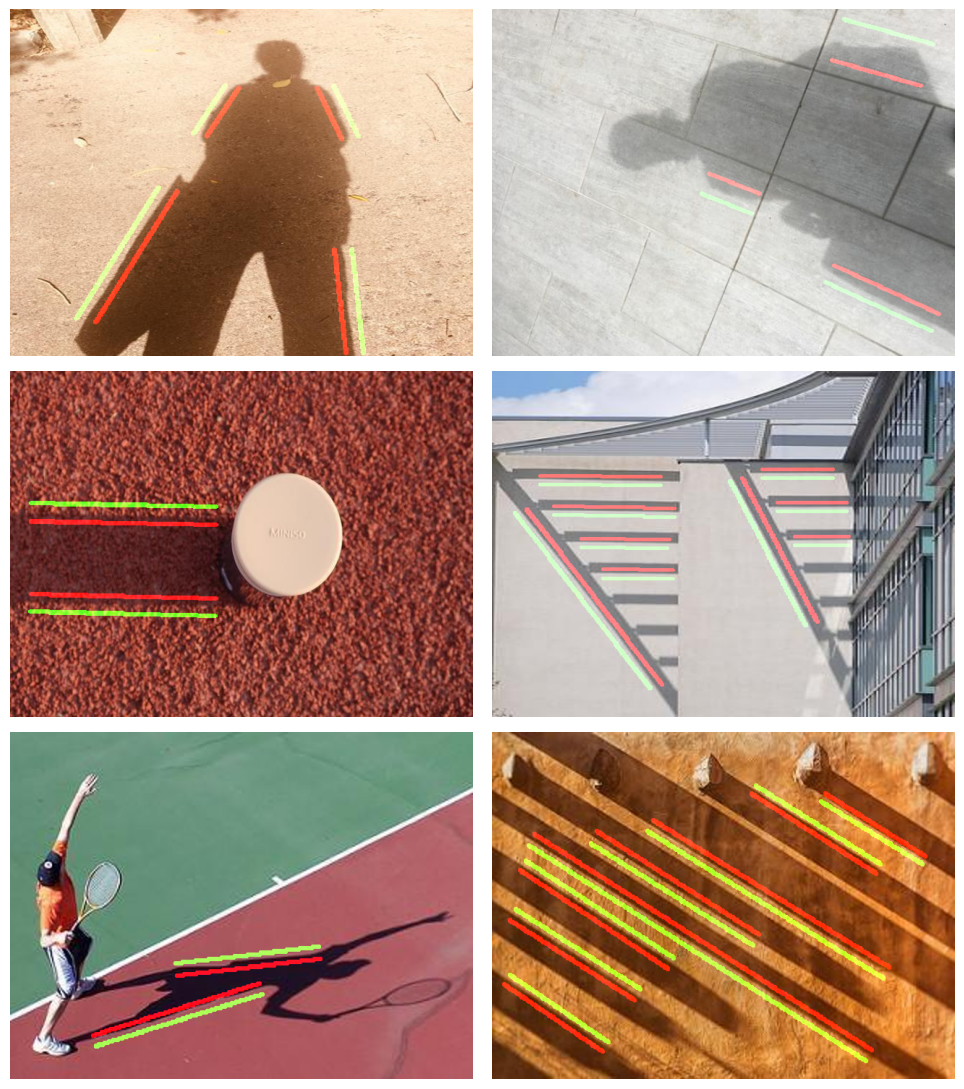}
    \caption{Shadow Images from CUHK-S}
  \end{subfigure}
  \caption{Example images from our proposed benchmark test set for shadow removal performance evaluation. Annotated pixels alongside the shadow edges are also shown per image, $S_{in}$ in red and $S_{out}$ in green. (a) shows examples from the SBU-S, and (b) shows images from the CUHK-S.}
  \label{fig:dataset}
  \vspace{-3mm}
\end{figure}

To address this gap, we curate a benchmark test set of shadow images in the wild. These images are sourced from existing shadow detection datasets, namely SBU \cite{vicente2016large} and CUHK \cite{hu2021revisiting}, exhibiting more complex scenes and larger variations in illumination, and thus better-representing shadow images in real-world scenarios. 
The proposed dataset contains 400 images in total. Among them, 210 shadow images are from SBU (SBU-S), and 190 images from CUHK (CUHK-S). 
To enable assessment of shadow removal performance on the proposed benchmark test set, we annotate pixels alongside the shadow edges that traverse the same material for each image in the dataset, examples of images and corresponding annotations are shown in \cref{fig:dataset}.
We believe that this dataset can serve as a valuable resource for evaluating shadow removal performance for general shadow images in real-world scenarios.

Another challenge in assessing shadow removal for shadow images in the wild is the lack of an evaluation metric. 
All current metrics used in shadow removal evaluation, \eg Mean Absolute Error (MAE), Peak Signal-to-Noise Ratio (PSNR), and Structural Similarity Index (SSIM), necessitate the comparison between the shadow-removed result and the ground truth shadow-free image. 
However, due to the complexity of real-world scenarios, obtaining the shadow-free version of the scene is impractical.

To tackle the absence of an effective evaluation metric, we propose a straightforward metric called Color Distribution Difference (CDD). 
CDD is grounded in the assumption that the area alongside the shadow edge maintains the same underlying texture.
The CDD metric calculates the Earth Mover’s Distance (EMD) between the color distribution histograms of pixels on both sides of the shadow edges.
\begin{equation}
    CDD = \textrm{EMD}([Hist(S), Hist(NS)])
\end{equation}
where $S, NS$ denotes the pixel sets in the shadow region and the non-shadow region, respectively.

The CDD quantifies the disparity in color distribution and is correlated with the quality of shadow removal performance. 
A lower mean CDD value indicates a better shadow-removal outcome, showcasing alignment between the colors on both sides of the shadow edge. 
A lower variance suggests more consistent performance across various shadow appearances, illustrating the method's generalizability.

\section{Experiments}
\noindent \textbf{Implementation Details.}
The proposed method is implemented using Pytorch, all experiments are tested on an NVIDIA TITAN RTX GPU.
We apply our method on top of two models, SP+M-Net \cite{Le_2019_ICCV} and ShadowFormer \cite{guo2023shadowformer}.
Hyper-parameters are set to be the same as in the training phase, except that the learning rate is set to $1e^{-5}$.
During the adaptation process, we update the pre-trained model for 20 iterations per image.
The refining parameters $\lambda_1, \lambda_2, \lambda_3$ and $\lambda_4$ are set to $1,1,0.1,10$ in our experiments. 
More details can be found in the supplementary.

\noindent \textbf{Datasets and Evaluation Metrics.}
We evaluate shadow removal performance for images-in-the-wild on our proposed test set, comprising 400 shadow images from SBU \cite{vicente2016large} and CUHK \cite{hu2021revisiting}, with material-consistent shadow edge pixels annotated per image. 
As shadow-free images are not available in our test set, we use the proposed Color Distribution Difference (CDD) as the quantitative evaluation metric. 
Additional evaluation results on the ISTD+ \cite{Le_2019_ICCV} dataset are provided in the supplementary materials.

\subsection{Comparison with SOTA Models}
To demonstrate the seamless integration of our proposed adaptation method with existing shadow removal models, we apply our method on top of three state-of-the-art (SOTA) models: SP+M-Net \cite{Le_2019_ICCV}, a CNN-based model, ShadowFormer \cite{guo2023shadowformer}, a Transformer-based model, and ShadowDiffusion \cite{guo2023shadowdiffusion}, a diffusion-based model.

\begin{table}[!t]
\centering
\caption{Comparison with SOTA models. We report the performance of existing state-of-the-art (SOTA) methods on our proposed test set and compare them with our adaptation method. CDD mean and variance values are provided, note that these values are reported as $1000\times$ the original value. $*$ denotes that only the SRD pre-trained model is available for evaluation.}
\label{table:cdd}
\resizebox{\linewidth}{!}
{
\begin{tabular}{@{}lcccc@{}}
\toprule
\multicolumn{1}{c}{\multirow{2}{*}{Methods}}  & \multicolumn{2}{c}{CUHK-S}      & \multicolumn{2}{c}{SBU-S}       \\ \cmidrule(l){2-5} 
\multicolumn{1}{c}{}                    & CDD Mean & CDD Var & CDD Mean & CDD Var \\ \midrule
Input                                   & 291.6    & 125.0   & 279.0    & 119.0   \\ \midrule
Inpaint4Shadow \cite{li2023leveraging}  & 25.3     & 36.0    & 28.6     & 40.7    \\ \midrule
ShadowDiffusion$^*$ \cite{guo2023shadowdiffusion} & 50.2 & 89.1 & 82.3  & 111.9   \\
ShadowDiffusion$^*$+Ours                & 48.7     & 77.1    & 78.2     & 103.4   \\ \midrule
SP+M-Net \cite{Le_2019_ICCV}            & 41.7     & 52.9    & 43.0     & 54.9    \\
SP+M-Net+Ours                           & 22.2     & 36.5    & 23.9     & 46.7    \\ \midrule
ShadowFormer \cite{guo2023shadowformer} & 23.3     & 33.6    & 27.9     & 45.4    \\ 
ShadowFormer+Ours                             & \textbf{16.4} & \textbf{31.9} & \textbf{15.0} & \textbf{26.3} \\ \bottomrule
\end{tabular}
}
\end{table}

\begin{figure*}[!t]
  \centering
  \includegraphics[width=\linewidth]{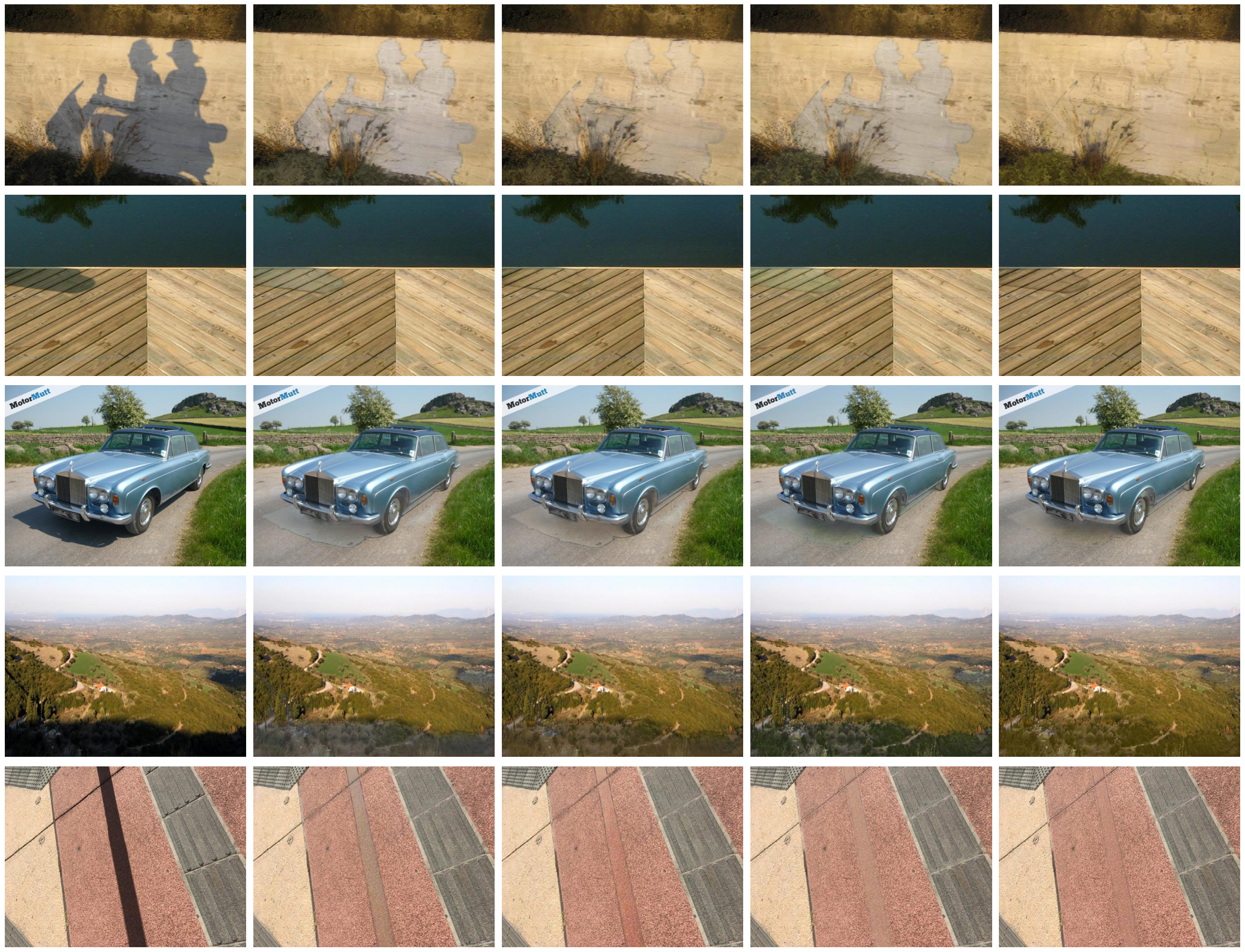}
  \makebox[0.19\linewidth]{\small{(a) Input}}
  \makebox[0.19\linewidth]{\small{(b) SP+M-Net \cite{Le_2019_ICCV}}}
  \makebox[0.19\linewidth]{\small{(c)\cite{Le_2019_ICCV}+Ours}}
  \makebox[0.19\linewidth]{\small{(d) ShadowFormer\cite{guo2023shadowformer}}}
  \makebox[0.19\linewidth]{\small{(e)\cite{guo2023shadowformer}+Ours}}
  \caption{Examples from test cases in our proposed dataset, (a) shows input images from our curated dataset, first three rows are from SBU-S and last two rows are from CUHK-S, (b) and (c) present pre-trained and refined SP+M-Net \cite{Le_2019_ICCV} results, (d) and (e) show pre-trained and refined ShadowFormer \cite{guo2023shadowformer} results.}
  \label{fig:qual}
\end{figure*}

\noindent \textbf{Quantitative evaluation.} For fair comparisons, we evaluate the performance of all existing methods using models trained on the ISTD+ dataset (except for ShadowDiffusion, where only the SRD-trained model is available). 
As shown in \cref{table:cdd}, our adaptation method enables both SP+M-Net and ShadowFormer to surpass their original results, with ShadowFormer achieving the best performance when our adaptation is applied. 
The high errors from existing SOTA methods indicate limited adaptability to various complex shadow scenes in the wild, as they are trained on simple and limited scenes. 
In contrast, our adaptation method leverages self-supervision from shadow edges in each image, refining each case during test time. 
Notably, the SRD-trained ShadowDiffusion performs the worst among the SOTA methods due to the intensity differences between the training shadow and shadow-free pairs, which we discuss further in \cref{sec:cross}.

\noindent \textbf{Qualitative evaluation.} As shown in \cref{fig:qual}, we demonstrate the improvement with our proposed adaptation method on both base models. Column (a) shows the input shadow images. In columns (b) and (d), we can see that pre-trained models struggle with complex shadows of various shapes, near the dark materials, and with occluders in the scene, resulting in obvious artifacts in the shadow region. With our adaptation method, as shown in columns (c) and (e), we fix the color shift in the shadow region. This is because our proposed adaptation constrains the color and texture consistency between shadow/non-shadow regions.

\subsection{Usefulness of CDD}
\begin{figure}[!t]
  \centering
  \includegraphics[width=0.65\linewidth]{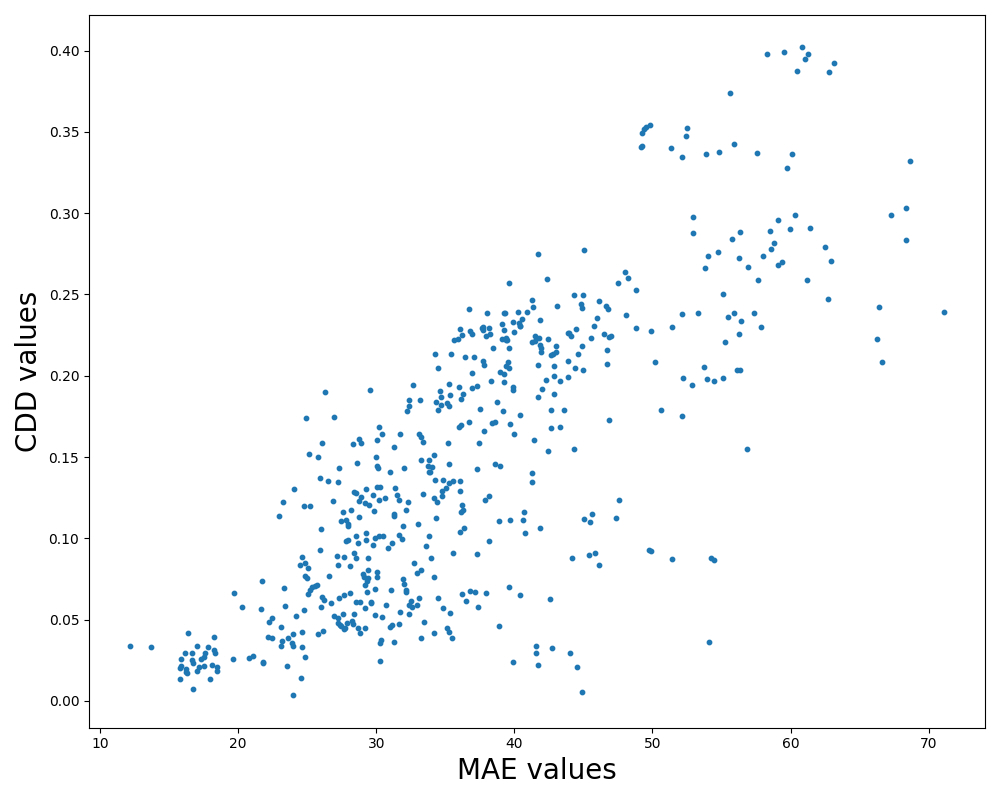}
  \caption{Per-image CDD and MAE errors for shadow images of the ISTD+ \cite{Le_2019_ICCV} test set. Each point represents the CDD and MAE error computed for a single image. The two metrics are highly correlated. Note that measuring MAE requires the paired shadow-free image while measuring our proposed CDD only requires annotating MC-shadow edges in the input shadow image.}
  \label{fig:maecdd}
\end{figure}

We propose the Color Distribution Difference (CDD) metric to address the lack of evaluation metrics for shadow removal in the wild, as existing metrics require ground truth shadow-free images. To validate the usefulness of our CDD metric, we plot the Mean Absolute Error (MAE) values alongside the CDD values for each shadow image in the ISTD+ test set \cite{Le_2019_ICCV} in \cref{fig:maecdd}. 
This plot reveals a strong correlation between the two metrics, demonstrating that CDD is a reliable alternative for evaluating shadow removal methods when shadow-free ground truth is not available.

\subsection{Ablation Studies}
\begin{figure}[!t]
   \centering
   \includegraphics[width=\linewidth]{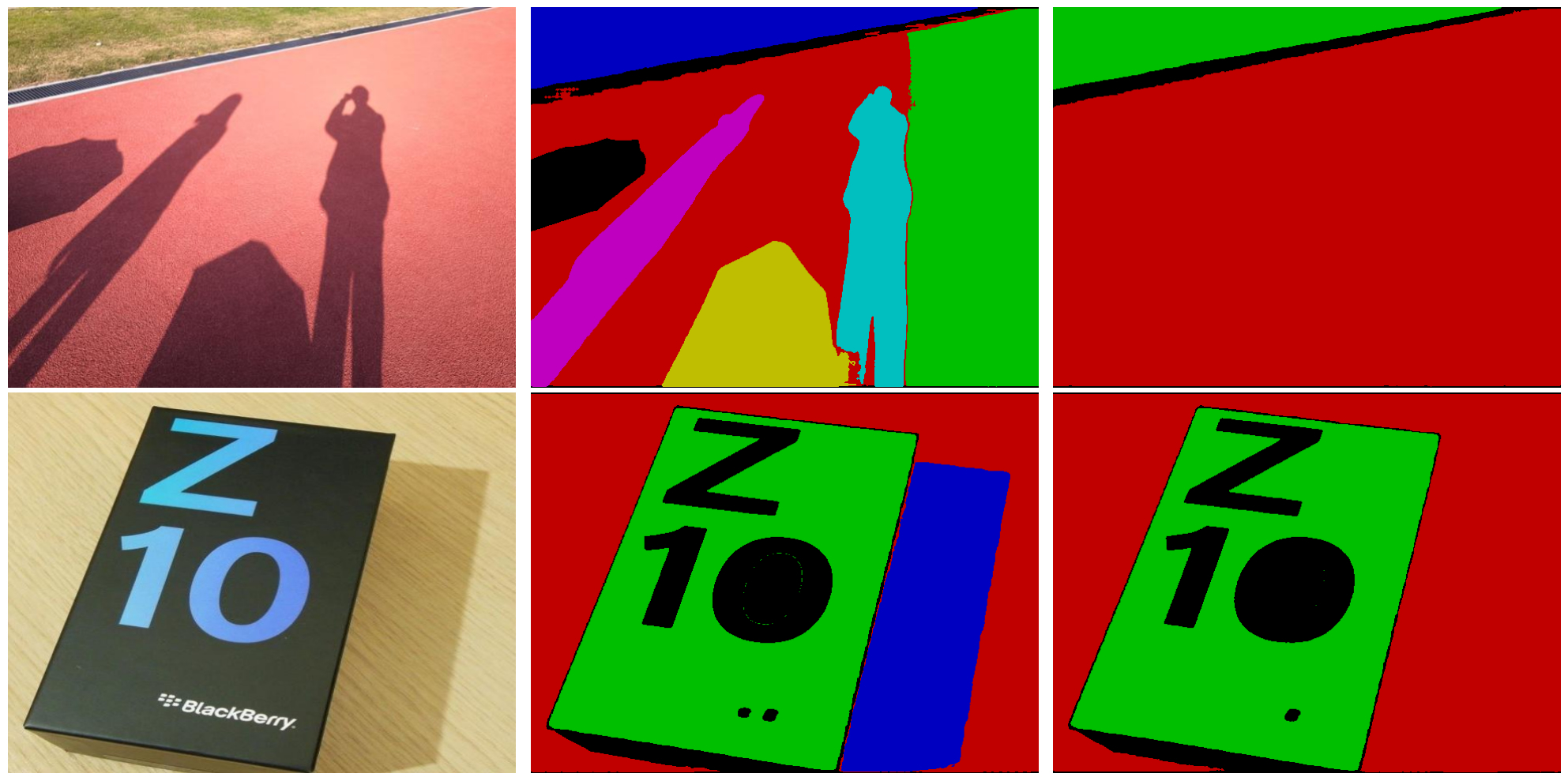}
    \makebox[0.32\linewidth]{\small{Input}}
    \makebox[0.32\linewidth]{\small{Vanilla SAM}}
    \makebox[0.32\linewidth]{\small{Fine-tuned SAM}}
   \caption{Comparison of vanilla SAM and fine-tuned SAM on test images. Our fine-tuned SAM predicts shadow-invariant masks while vanilla SAM is sensitive to shadow regions.}
   \label{fig:comparesam}
   \vspace{-3mm}
\end{figure}

\noindent \textbf{The effect of Fine-tuned SAM.} 
We fine-tune the Segment Anything Model \cite{kirillov2023segany} to extract shadow edges that traverse the same material. In \cref{fig:comparesam}, we demonstrate that the fine-tuned SAM produces shadow-invariant masks.
Additionally, \cref{table:ablation} compares the performance of using vanilla SAM and our fine-tuned SAM for adaptation. The latter surpasses the former due to its reduced sensitivity to shadow presence.

\begin{figure}[!t]
  \centering
  \includegraphics[width=\linewidth]{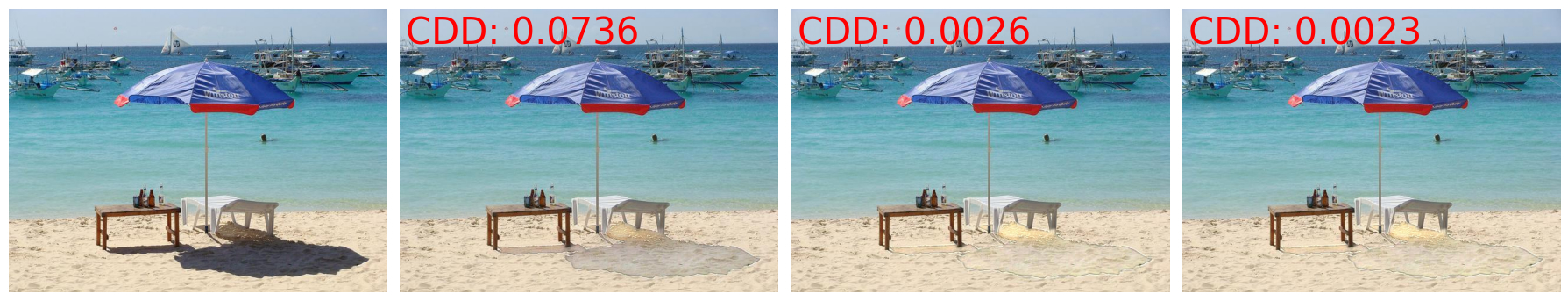}
  \makebox[0.24\linewidth]{\small{Input}}
  \makebox[0.24\linewidth]{\small{Per Mask}}
  \makebox[0.24\linewidth]{\small{Pixels}}
  \makebox[0.24\linewidth]{\small{Pixels\&Patches}}
  \caption{Comparison of different refinement configurations. Refining pixels on all extracted shadow edges and patches per material mask yields the best performance.}
  \label{fig:compareconfig}
\end{figure}

\begin{figure}[!t]
  \centering
  \includegraphics[width=\linewidth]{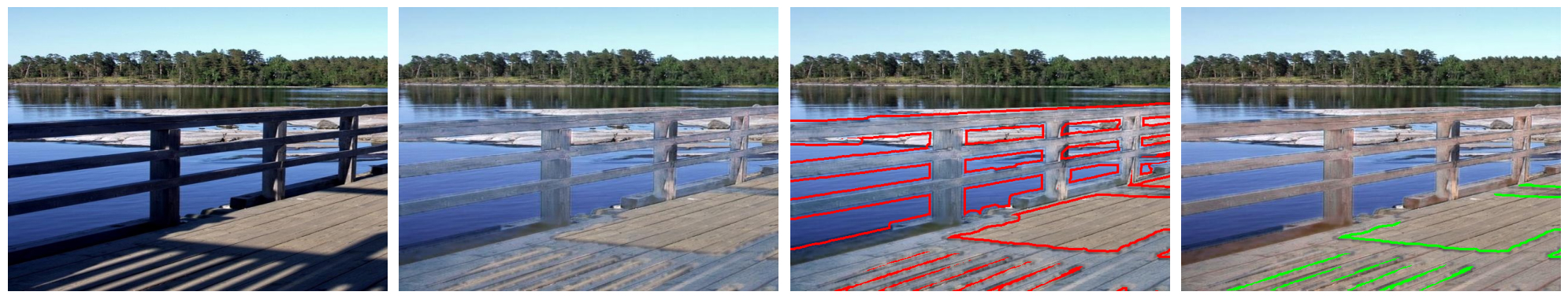}
  \makebox[0.24\linewidth]{\small{Input}}
  \makebox[0.24\linewidth]{\small{\cite{guo2023shadowformer}}}
  \makebox[0.24\linewidth]{\small{All Edges}}
  \makebox[0.24\linewidth]{\small{MC Edges}}
  \caption{Example of using proposed edge extraction. We show the refined outputs using all shadow edges (colored in red), and the refined outputs using only our extracted material-consistent shadow edges (colored in green). Using extracted MC edges improves the performance from 0.0372 to 0.0039.}
  \label{fig:edge}
  \vspace{-3mm}
\end{figure}

\noindent \textbf{The effectiveness of the Proposed Adaptation Method.}
\cref{table:ablation} shows evaluations of different configurations for the adaptation method. We find that the proposed approach, which combines fine-tuned SAM edge extraction with pixel and patch sampling for supervision, yields the overall best results.
``Per Mask" refines the edge pixels and patches on each SAM-detected material mask that intersects with the shadow mask, and ``Pixels" refines only the pixels on all extracted edges at once. We can see that the ``Pixels\&Patches" configuration achieves the best overall results (shown in \cref{fig:compareconfig}), as it maintains texture consistency within each material while ensuring global consistency across all the extracted shadow edges. When no edge extraction is performed, the entire shadow edge is used, including those edges coinciding with material boundaries, the adaptation process is misguided because these edges separate different materials on either side, as depicted in \cref{fig:edge}.

\begin{table}[!t]
\centering
\caption{Quantitative results with different configurations of the adaptation method and with different loss settings. We report the CDD mean and variance values on our proposed dataset. Note that the CDD values are reported in $1000\times$ the original value.}
\label{table:ablation}
\resizebox{\columnwidth}{!}{%
\begin{tabular}{@{}lcccc@{}}
\toprule
\multicolumn{1}{c}{}                        & \multicolumn{2}{c}{CUHK}      & \multicolumn{2}{c}{SBU}       \\ \midrule
\multicolumn{1}{c}{\textbf{Configurations}} & CDD Mean      & CDD Var       & CDD Mean      & CDD Var       \\ \midrule
VanillaSAM+Pixels\&Patches                  & 19.6          & 34.4          & 21.3          & 44.9          \\ \midrule
FinetunedSAM+Per Mask                       & 16.9          & \textbf{29.3} & 16.8          & 33.6          \\ \midrule
FinetunedSAM+Pixels                         & 16.9          & 32.1          & 15.5          & 27.0          \\ \midrule
FinetunedSAM+Pixels\&Patches                & \textbf{16.4} & 31.9          & \textbf{15.0} & \textbf{26.3} \\ \midrule
\multicolumn{1}{c}{\textbf{Losses}}         & CDD Mean      & CDD Var       & CDD Mean      & CDD Var       \\ \midrule
All losses                                  & \textbf{16.4} & \textbf{31.9} & \textbf{15.0} & 26.3          \\ \midrule
- $L_{distance}$                            & 16.8          & 32.7          & 15.0          & \textbf{25.8} \\ \midrule
- $L_{distribution}$                        & 19.2          & 30.9          & 21.4          & 43.6          \\ \midrule
- $L_{per}$                                 & 16.9          & 32.1          & 15.5          & 27.0          \\ \bottomrule
\end{tabular}%
}
\end{table}

\noindent \textbf{Proposed Losses.}
Our adaptation method comprises four loss functions.
In \cref{table:ablation},  we present experimental results by ablation of each of these losses.
$L_{distance}$ and $L_{distribution}$ enforce color matching and distribution alignment along the shadow edges, while the $L_{per}$ ensures texture consistency within the same material. The combination of these three losses achieves the best performance.


\subsection{Cross dataset testing}
\label{sec:cross}
To further demonstrate our adaptation method's performance on adapting to out-of-distribution images, we apply our method on ShadowFormer \cite{guo2023shadowformer} pre-trained on the SRD dataset \cite{qu2017deshadownet} (shadow masks provided by DHAN \cite{cun2020towards}) and the ISTD dataset \cite{wang2018stacked} to test on the ISTD+ test set.
Note that images from the SRD and ISTD datasets exhibit different light intensities between training shadow and shadow-free image pairs, due to the image acquisition process. Thus, models pre-trained on those datasets often alter the overall color intensity of the whole input image.

\begin{figure}[!t]
  \centering
  \includegraphics[width=\linewidth]{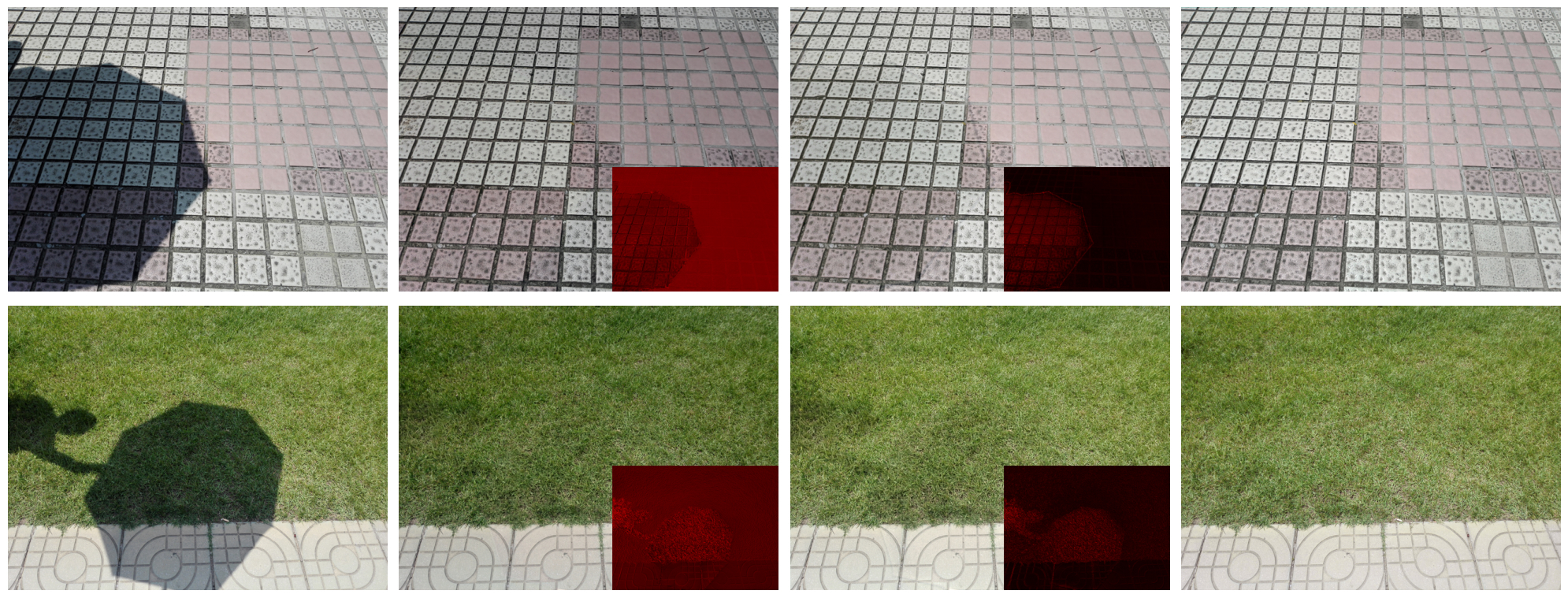}
  \makebox[0.24\linewidth]{\small{(a) Input}}
  \makebox[0.24\linewidth]{\small{(b) \cite{guo2023shadowformer}}}
  \makebox[0.24\linewidth]{\small{(c) \cite{guo2023shadowformer}+Ours}}
  \makebox[0.24\linewidth]{\small{(d) GT}}
  \caption{Qualitative comparison on cross dataset testing. We use ISTD pre-trained ShadowFormer and test it on the ISTD+ test set. (a) shows input image, (b) shows ShadowFormer \cite{guo2023shadowformer} result, (c) presents the results with refinement, and (d) presents the ground truth. We also plot the error maps on the corner.}
  \label{fig: cross}
\end{figure}

To address this, we additionally calculate the Mean Squared Error (MSE) loss of the non-shadow region pixels, denoted as $L_{nonshadow}$.
\cref{table: cross} presents the ISTD+ test results of those models with and without our proposed adaptation method. With our adaptation method, the output images exhibit correct colors for both shadow and non-shadow regions, outperforming pre-trained models in both MAE and CDD measurements.
As illustrated in \cref{fig: cross}, the ISTD pre-trained model alters the overall light intensity of the image, resulting in subpar performance on ISTD+ test cases. In contrast, our adaptation method effectively corrects this error through the design of $L_{nonshadow}$, which enforces color consistency in the non-shadow regions. 

\begin{table}[!t]
\centering
\caption{Quantitative results on cross dataset testing. ISTD pre-trained ShadowFormer and SRD pre-trained ShadowFormer are tested on the ISTD+ test set. }
\resizebox{\linewidth}{!}
{
\begin{tabular}{@{}cclccccc@{}}
\toprule
\multicolumn{1}{c}{\multirow{2}{*}{Trained On}} &
  \multicolumn{1}{c}{\multirow{2}{*}{Tested On}} &
  \multirow{2}{*}{Methods} &
  \multicolumn{3}{c}{MAE} &
  \multicolumn{2}{c}{CDD} \\ \cmidrule(l){4-8} 
\multicolumn{1}{c}{} &
  \multicolumn{1}{c}{} &
   &
  \multicolumn{1}{c}{S} &
  \multicolumn{1}{c}{NS} &
  \multicolumn{1}{c}{A} &
  \multicolumn{1}{c}{Mean} &
  \multicolumn{1}{c}{Var} \\ \midrule
\multirow{2}{*}{SRD}  & \multirow{2}{*}{ISTD+} & w.o. Ours & 13.7 & 3.4 & 5.1 & 55.0 & 43.3 \\ \cmidrule(l){3-8} 
                      &                        & w. Ours   & 6.2  & 2.4 & 3.0 &  8.0 & 9.4  \\ \midrule
\multirow{2}{*}{ISTD} & \multirow{2}{*}{ISTD+} & w.o. Ours & 10.6 & 6.3 & 7.0 & 11.8 & 17.7 \\ \cmidrule(l){3-8} 
                      &                        & w. Ours   & 6.3  & 2.7 & 3.4 &  1.0 & 3.1  \\ \bottomrule
\end{tabular}
}
\label{table: cross}
\end{table}

\subsection{Limitations}
\begin{figure}[!t]
   \centering
   \includegraphics[width=\linewidth]{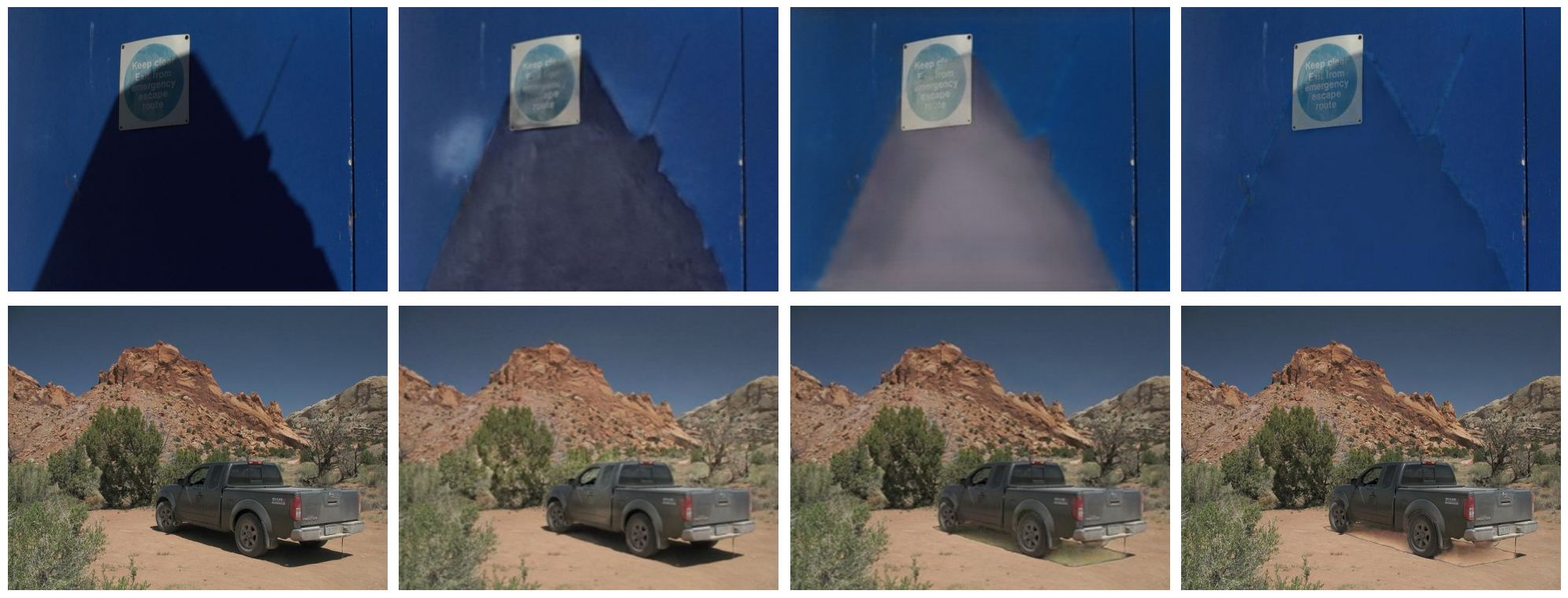}
   \makebox[0.24\linewidth]{\small{(a) Input}}
   \makebox[0.24\linewidth]{\small{(b) \cite{guo2023shadowdiffusion}}}
   \makebox[0.24\linewidth]{\small{(c) \cite{li2023leveraging}}}
   \makebox[0.24\linewidth]{\small{(d) Ours}}
   \caption{\textbf{Limitation on shadow edges.} We show the results of SOTA methods on our test images, visible edge artifacts are caused by the limited adaptability of pre-trained models. Our refinement mainly focuses on the color discrepancy in the shadow region.}
   \label{fig: fail}
\end{figure}

Our adaptation method has several limitations that can be interesting directions for future work. 
First, we do not enforce specific constraints on the penumbra shadow regions. Handling the smooth changing shadow effects on those areas is challenging for all shadow removal methods, including ours, which often leaves noticeable shadow edge artifacts (see \cref{fig: fail}). While our method focuses on fixing the color discrepancy, future research could explore mitigating edge artifacts, particularly for out-of-distribution test cases.
Second, our approach involves iterative refinement of the pre-trained model using extracted self-supervision, leading to a computational overhead of approximately 24 seconds per image. Last, we mainly focus on the supervision obtained from material-consistent edges. The remaining part of the shadow edge might also be helpful for understanding and removing the shadow. 


\section{Conclusion}
We introduce a test-time self-supervised adaptation method for deep-learning shadow removal.
To gather the supervision signal, we fine-tune the image foundation model, SAM, to generate shadow-invariant segmentation masks, effectively extracting shadow edges that traverse the same material. 
Pixels near these edges and patches within the same material provide valuable supervision.
We then propose an iterative adaptation approach for the pre-trained model using the collected supervision to ensure color and texture consistency.
We demonstrate that our proposed losses significantly enhance deep shadow removal, both qualitatively and quantitatively, across various challenging testing cases.
Moreover, we introduce a benchmark test set and a metric that enable the evaluation of shadow removal methods on images with complex shadows, even in the absence of shadow-free ground truth images.





{\small
\bibliographystyle{ieee_fullname}
\bibliography{egbib,hthesis}
}

\clearpage
\setcounter{section}{0}
\setcounter{figure}{0}
\setcounter{table}{0}

\noindent In this supplementary material, we provide the following:
\begin{enumerate}
    \item Details of material-consistent shadow edge extraction.
    \item Correctness of the proposed CDD metric.
    \item Using shadow edges from shadow detection models.
    \item Further details of implementation.
    \item Quantitative results on ISTD+ test set.
    \item More results on the cross-dataset testing.
    \item More qualitative results.
\end{enumerate}

\section{Details of Material-Consistent Shadow Edge Extraction}

\begin{figure}[!ht]
  \centering
  \includegraphics[width=\linewidth]{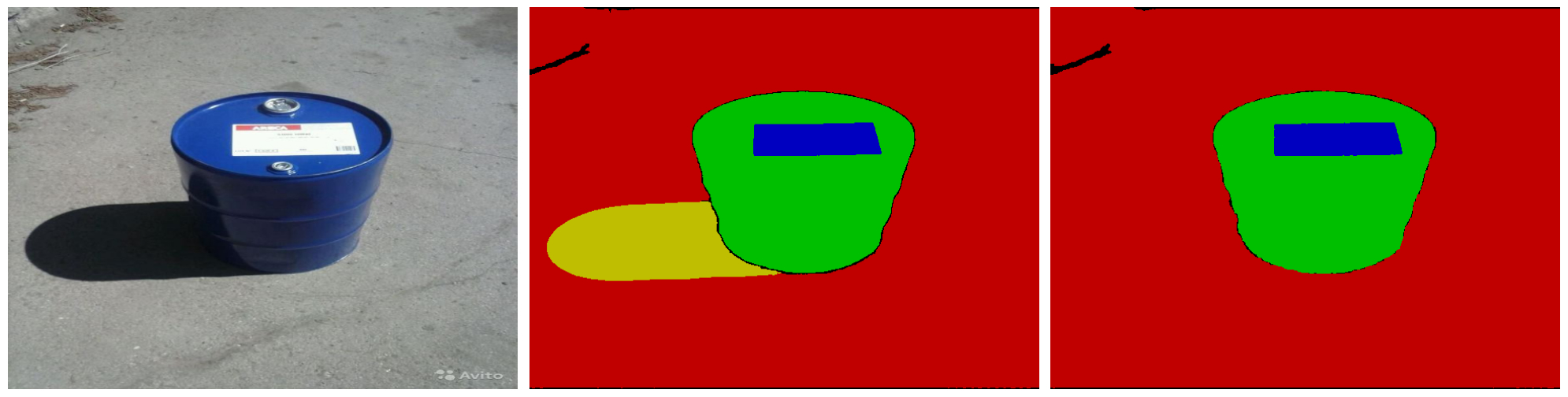}\\
  \includegraphics[width=\linewidth]{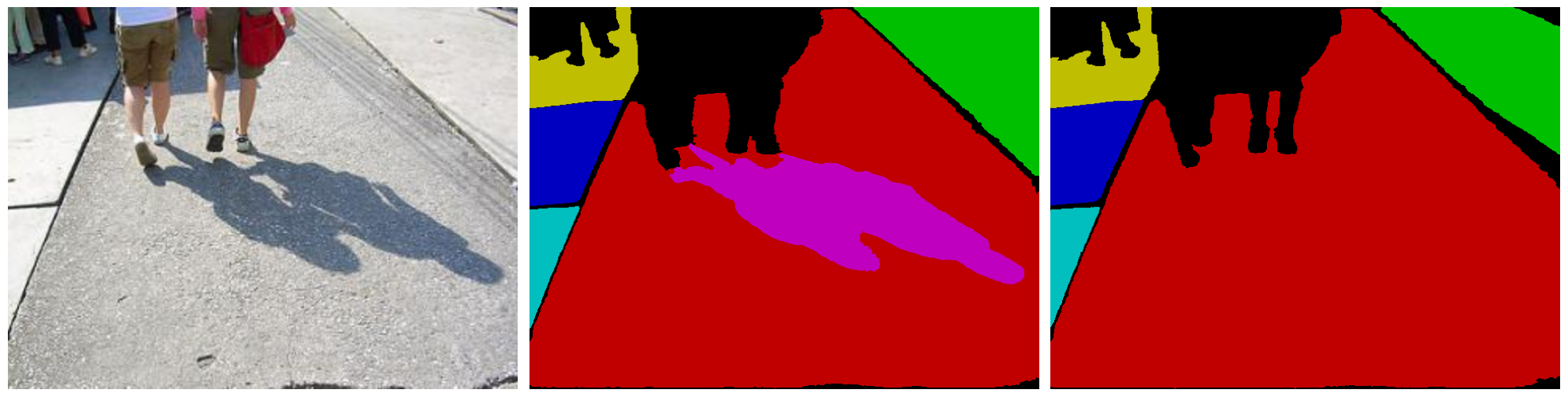}
  \makebox[0.32\linewidth]{\small{Input}}
  \makebox[0.32\linewidth]{\small{Vanilla SAM}}
  \makebox[0.32\linewidth]{\small{Fine-tuned SAM}}
  \caption{More examples of improved material-consistent mask segmentation by our fine-tuned SAM.}
  \label{fig: material mask}
\end{figure}

We select only material-consistent shadow edges (MC-Edges) and enforce color and texture consistency on both sides of these edges on the shadow-removed outputs. 
These constraints should not be enforced for shadow edges aligning with object boundaries, as both sides of those edges should exhibit different shadow-free textures and colors. 
To extract MC edges, we first use the provided $SamAutomaticMaskGenerator$ function from the Segment Anything Model (SAM) \cite{kirillov2023segany} to predict material masks, and then sample edge pixels and patches when the material masks intersect with the shadow mask, details of the process is described in \cref{algo: extract}. More visual examples of improved material mask segmentation by our fine-tuned SAM are shown in \cref{fig: material mask}.

\begin{algorithm}[!ht]
\SetAlgoVlined
\KwData{Input shadow image $I$, shadow mask $M$; \textbf{Model:} Fine-tuned SAM $f_{SAM}$}
\KwResult{Sampled shadow/shadow-free pairs, $Pixel_{in/out}$ and $Patch_{in/out}$}
 $EdgePixels$ = $\{M-erode(M)\}$ + $\{dilate(M)-M\}$ \\
 $SegMasks$ = $MaskGenerator$($f_{SAM}$, $I$, $M$) \\
 \For{$i = 1 \rightarrow n$}{
    \uIf{$SegMask[i]$ overlaps with $M$}{
    Get $Pixel_{in/out}$ in $EdgePixels \in SegMask[i]$  \\
    Get $Patch_{in/out}$ in $SegMask[i]$ \;
    }
    \Else{
    continue \;
  }
 } 
\caption{Material-Consistent Shadow Edge Extraction}
\label{algo: extract}
\end{algorithm}

To demonstrate the effectiveness of fine-tuned SAM in extracting material-consistent shadow edge, we compare its performance to vanilla SAM on different materials. \cref{tab: per scene} presents the percentage of edge pixels extracted by each SAM model across various materials in the ISTD+ dataset \cite{Le_2019_ICCV}. The results indicate that our proposed method significantly outperforms vanilla SAM, successfully extracting shadow edge pixels across different material types.

\begin{table}[!ht]
\centering
\caption{Percentage of material-consistent shadow edge pixels detected (recall\%) via vanilla SAM and our fine-tuned SAM.}
\label{tab: per scene}
\resizebox{0.7\columnwidth}{!}{%
\begin{tabular}{@{}lcc@{}}
\toprule
\multicolumn{1}{c}{Material Type} & Vanilla SAM & Fine-tuned SAM \\ \midrule
grass                             & 85.7        & 98.6           \\
cement                            & 63.6        & 96.0           \\
ceramic                           & 66.1        & 71.0           \\
playground                        & 75.6        & 82.8           \\ \bottomrule
\end{tabular}%
}
\end{table}
\section{Correctness of the Proposed CDD Metric}

\begin{figure}[!ht]
  \centering
  \includegraphics[width=\linewidth]{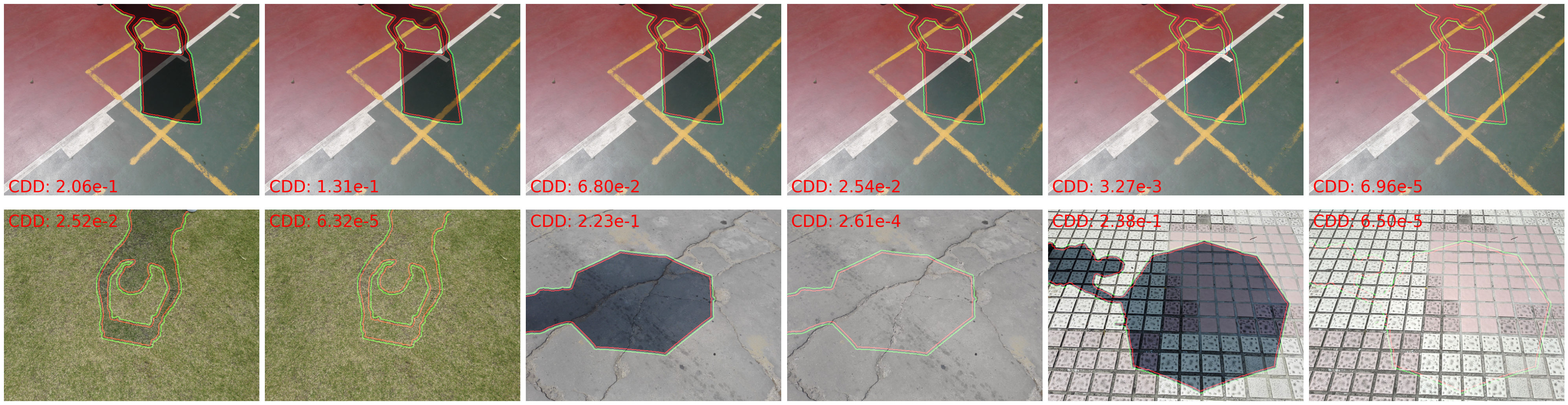}
  \caption{Correctness of Color Distribution Difference metric. (\textit{top}) We manually adjust the shadow intensity from strong to weak, and the CDD values are lower when the shadow effect is weaker. (\textit{bottom}) We compare the CDD values of shadow images and their shadow-free counterparts, the CDD value of the shadow-free version is at least two orders of magnitude lower than the shadow version. CDD are computed using the pixels marked in the images and the values are reported in the images.}
  \label{fig: cddeffect}
\end{figure}

To justify the correctness of our proposed Color Distribution Difference (CDD) metric, we first show that the CDD value corresponds to the shadow intensity. As shown in the top row of \cref{fig: cddeffect}, we choose a shadow image and manually adjust the shadow intensity. We find that the weaker the shadow effect, the lower the CDD results, indicating that the CDD metric can effectively represent the quality of shadow removal.

Additionally, we show the CDD results on ground truth images from ISTD+ \cite{Le_2019_ICCV}. We compare the CDD values of the shadow-free images against their shadow counterparts, as shown in the bottom row of \cref{fig: cddeffect}. The CDD values for shadow-free images are at least two orders of magnitude lower than those for shadow images. Therefore, we believe CDD serves as a valuable metric for evaluating shadow removal performance when ground truth is unavailable.

\begin{figure}[!ht]
  \centering
  \includegraphics[width=\linewidth]{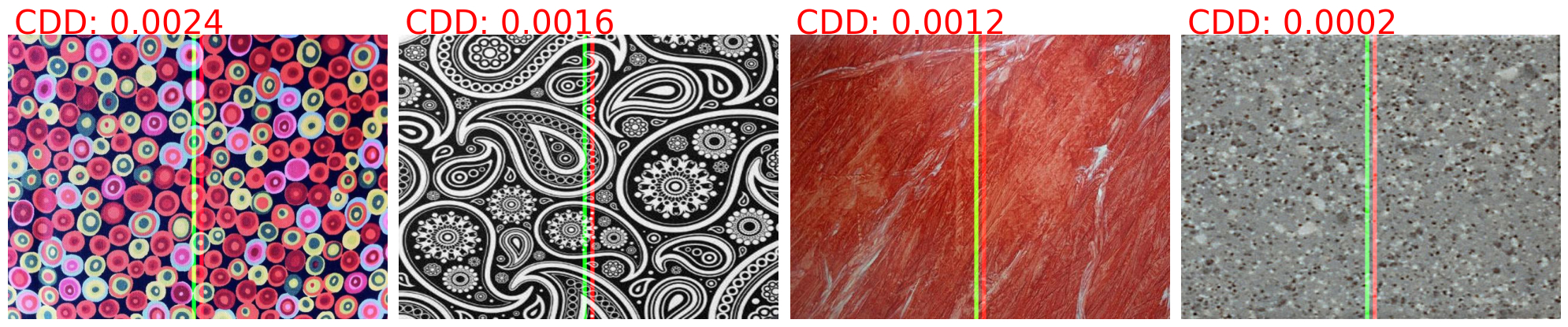}
  \caption{Validation of our proposed CDD metric on DTD \cite{cimpoi14describing}. We show examples from the subset. We apply a simple method to compute the CDD values on the DTD data using parallel lines in the middle (see the $red$ and $green$ lines in each image). We can see that this simple annotation still yields low CDD scores.}
  \label{fig: dtd}
\end{figure}

Finally, to validate the adequacy of our proposed CDD metric in measuring consistency alongside shadow edges, we select a subset from the DTD dataset \cite{cimpoi14describing}, which we believe showcases more complex textures than our proposed shadow image dataset (\cref{fig: dtd}). 
We evaluate the CDD score on pixels from parallel lines in the middle of the image. This result can be considered the upper bound of the ground truth CDD score for our proposed dataset. 
The CDD measurement on this selected subset is 0.0022, significantly lower than the shadow removal results in the main paper (0.0157). This evaluation further demonstrates that the proposed CDD metric can effectively serve as a valuable shadow removal evaluation metric.

\section{Shadow-Removal Refinement Results Using Shadow Edges From Shadow Detection Models}

In our main paper, we conduct experiments using ground truth shadow masks sourced from established datasets \cite{Le_2019_ICCV,vicente2016large,hu2021revisiting}. 
These masks might not be available for ideal automated shadow removal systems used in real-world scenarios.
In this section, we show that our method can be used with shadow masks detected from a shadow detection method. We must note that shadow detection is a relatively easier task compared to shadow removal and a robust, scale-able shadow detection system is more feasible since shadow detection training data is much easier to obtain. 
At some point, one can expect to get accurate shadow masks automatically, which could be directly incorporated into our system to improve shadow removal. 

\begin{figure}[!t]
  \centering
  \includegraphics[width=\linewidth]{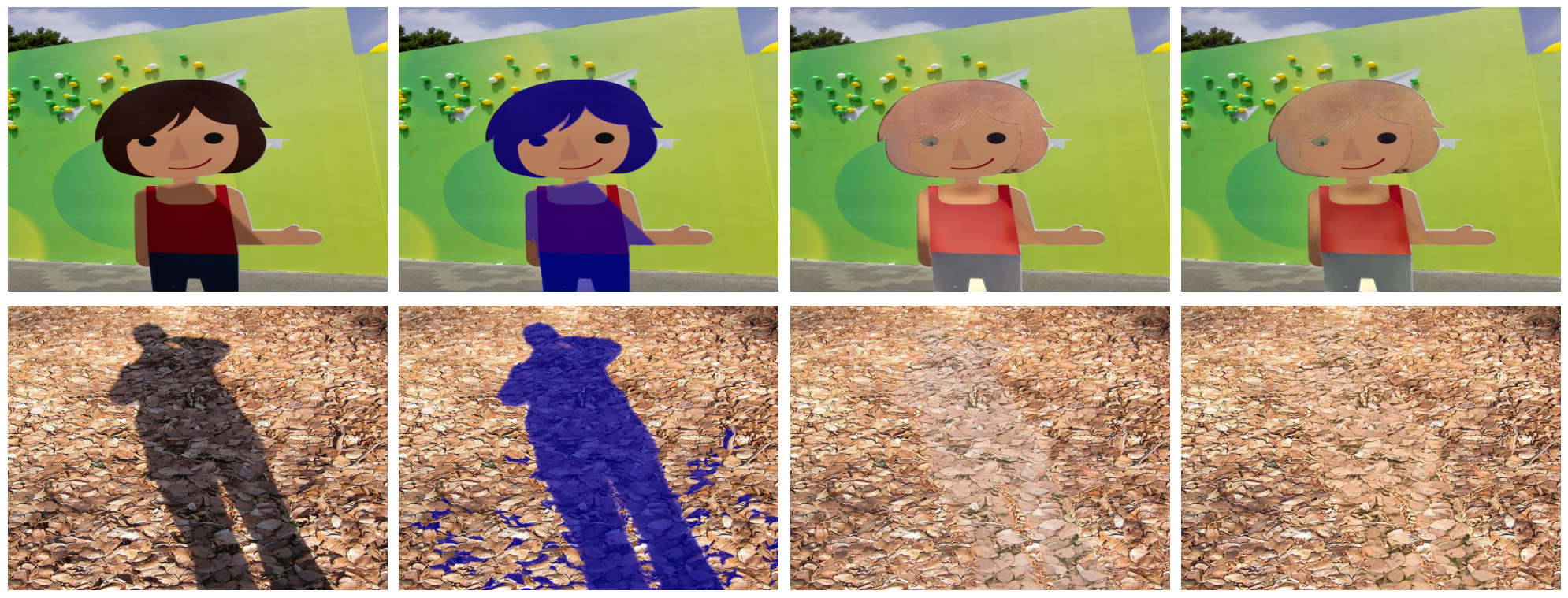}
  \makebox[0.24\linewidth]{\small{Input}}
  \makebox[0.24\linewidth]{\small{Detected mask}}
  \makebox[0.24\linewidth]{\small{\cite{guo2023shadowformer}}}
  \makebox[0.24\linewidth]{\small{Ours}}
  \caption{Examples of using SILT \cite{yang2023silt} detected masks for shadow removal refinement. Top row shows a failure case in ISTD+ \cite{Le_2019_ICCV}, where the dark region is predicted as the shadow region, leading to an inaccurate shadow removal result that our refinement method cannot rectify. Bottom row shows a successful case in the proposed dataset, where an accurate shadow mask is predicted and our refinement method improves the shadow removal performance.}
  \label{fig: detect}
\end{figure}

We use the state-of-the-art shadow detection method, SILT \cite{yang2023silt}, to generate shadow masks for each testing image. \cref{table: ber} presents the detection performance on the two testing datasets using SILT. 
Then, we compare the performance of ShadowFormer \cite{guo2023shadowformer} using ground truth shadow masks and these detected shadow masks in \cref{table: mask}. 
We find that on both the ISTD+ dataset \cite{Le_2019_ICCV} and our proposed dataset, using detected masks results in worse shadow-removal performance compared to using ground truth masks. A typical failure case is shown in the top row of \cref{fig: detect}.
Nevertheless, applying our method atop ShadowFormer \cite{guo2023shadowformer} can improve the performance in both cases, as shown in \cref{table: mask}. The bottom row of \cref{fig: detect} visualizes an example of how our method improves the shadow-removal result. 

\begin{table}[!t]
\centering
\caption{Shadow detection results on the ISTD+ \cite{Le_2019_ICCV} and our proposed dataset by SILT \cite{yang2023silt}. Following \cite{yang2023silt}, the performance is evaluated by the Balanced Error Rate ($BER$).}
\label{table: ber}
\begin{tabular}{@{}lccc@{}}
\toprule
Dataset  & $BER$ & $BER_S$ & $BER_{NS}$ \\ \midrule
ISTD+    & 1.12 & 0.80   & 1.44    \\ \midrule
Proposed & 4.05 & 4.02   & 4.09    \\ \bottomrule
\end{tabular}

\end{table}

\begin{table}[!t]
\centering
\caption{Quantitative comparison of results using ground truth shadow mask and SILT-detected shadow mask on \cite{guo2023shadowformer} and \cite{guo2023shadowformer}+Ours. MAE and CDD are reported. Note that CDD is reported in 1000× the original value.}
\label{table: mask}
\resizebox{\linewidth}{!}
{
\begin{tabular}{@{}lcc|ccccc@{}}
\toprule
\multirow{3}{*}{Methods}                    & \multicolumn{2}{c|}{Proposed} & \multicolumn{5}{c}{ISTD+}                         \\ \cmidrule(l){2-8} 
                                            & \multicolumn{2}{c|}{CDD}      & \multicolumn{3}{c}{MAE} & \multicolumn{2}{c}{CDD} \\ \cmidrule(l){2-8} 
                                            & Mean          & Var           & S      & NS     & A     & Mean        & Var       \\ \midrule
\cite{guo2023shadowformer} w. GT mask       & 25.0          & 40.8          & 5.3    & 2.2    & 2.7   & 1.5         & 2.6       \\ \midrule
Ours w. GT mask                             & 15.9          & 30.2          & 5.0    & 2.2    & 2.7   & 1.0         & 1.6       \\ \midrule
\cite{guo2023shadowformer} w. detected mask & 25.8          & 42.4          & 6.2    & 2.6    & 3.1   & 2.7         & 9.8       \\ \midrule
Ours w. detected mask                       & 19.3          & 38.5          & 5.7    & 2.5    & 3.0   & 2.4         & 8.4       \\ \bottomrule
\end{tabular}
}
\end{table}

\section{Further Details of Implementation}

\subsection{Computation Overhead}
Our approach iteratively refines the pre-trained model using extracted self-supervision. During testing on our proposed dataset, we update the whole model for 20 iterations per image, which results in an average of 24 seconds overhead on an NVIDIA TITAN RTX GPU. 

We further investigate the refining performance on different numbers of iterations and different model update policies (\eg updating only the last decoder layer of ShadowFormer \cite{guo2023shadowformer}). 
The results are shown in \cref{table: overhead}. Breaking down the overhead, our shadow edge extraction process takes 2 seconds, and each model update iteration requires approximately 1 second.
It is evident that the number of iterations is the primary cause of our computational overhead. Additionally, we found that partially updating the model does not improve the efficiency of our refinement and leads to decreased performance.

\begin{table}[!ht]
\centering
\caption{Comparison of performance and computation overhead using different numbers of iterations and update policies. We report the CDD mean values on the full proposed test set.}
\label{table: overhead}
\begin{tabular}{@{}cccc@{}}
\toprule
\# of Iter          & \multicolumn{1}{l}{Update} & CDD Mean & Overhead (s) \\ \midrule
10                  & whole                      & 17.6     & 13           \\ \midrule
\multirow{2}{*}{20} & whole                      & 15.7     & 24           \\ \cmidrule(l){2-4} 
                    & partial                    & 19.1     & 24           \\ \midrule
30                  & whole                      & 18.1     & 36           \\ \bottomrule
\end{tabular}
\end{table}

\subsection{Number and size of sampled patches}
In the main paper, we sample 8 patches of size $16 \times 16$ in our final configuration. Here, we experiment with different numbers and sizes of patches and compare the shadow removal performance on our proposed dataset. As shown in \cref{table: patch}, these hyper-parameters do not significantly affect overall performance.

\begin{table}[!t]
\centering
\caption{Quantitative comparison of using different numbers and sizes of sampled patches in our final configuration. Note that CDD values are reported in $1000\times$ the original value.}
\label{table: patch}
\resizebox{\linewidth}{!}
{
\begin{tabular}{|c|ccc|}
\hline
\diagbox{Number}{CDD Mean/Std.}{Size} & $8\times8$ & $16\times16$ & $32\times32$ \\
\hline
4   & 15.7/29.1 & 15.7/29.1 & 15.7/29.1 \\
\hline
8   & 15.7/29.2 & 15.7/29.1 & 15.7/29.0 \\
\hline 
16  & 15.8/29.2 & 15.7/29.1 & 15.7/29.1 \\
\hline
\end{tabular}
}
\end{table}

\subsection{Test settings}
We provide detailed hyper-parameter settings in SAM \cite{kirillov2023segany}, SID \cite{Le_2019_ICCV}, and ShadowFormer \cite{guo2023shadowformer} in \cref{table: setting}. 
For SAM, we use the $vit\_b$ model with the learning rate set to $1e^{-4}$. During fine-tuning, we extract 32 points per image as prompts. Enabling $multimask\_output=True$ for the mask decoder, we calculate the mean Dice loss \cite{sudre2017generalised} over three output masks per prompt.

\begin{table}[!t]
\centering
\caption{Detailed hyper-parameter settings used in SAM\cite{kirillov2023segany}, SID\cite{Le_2019_ICCV}, and ShadowFormer\cite{guo2023shadowformer}.}
\label{table: setting}
\resizebox{\linewidth}{!}
{
\begin{tabular}{@{}lc|lc@{}}
\toprule
\multicolumn{1}{c}{Hyper-parameter} & Value     & \multicolumn{1}{c}{Hyper-parameter} & Value     \\ \midrule
\multicolumn{2}{c|}{\textbf{SAM}}              & \multicolumn{2}{c}{\textbf{ShadowFormer}}      \\ \midrule
point\_per\_side                   & 16        & input\_size                        & (640,480) \\ \midrule
predict\_iou\_thres                & 0.90      & train\_ps                          & 320       \\ \midrule
stability\_score\_thres            & 0.90      & embed\_dim                         & 32        \\ \midrule
min\_mask\_region\_area            & 500       & win\_size                          & 10        \\ \midrule
\multicolumn{2}{c|}{\textbf{SID}}              & token\_projection                  & $linear$  \\ \midrule
input\_size                        & (512,512) & token\_mlp                         & $leff$    \\ \bottomrule
\end{tabular}
}
\end{table}

\subsection{MC-Edge Annotations for the ISTD+}
\begin{figure}[!ht]
  \centering
  \begin{subfigure}{0.48\linewidth}
    \includegraphics[width=\linewidth]{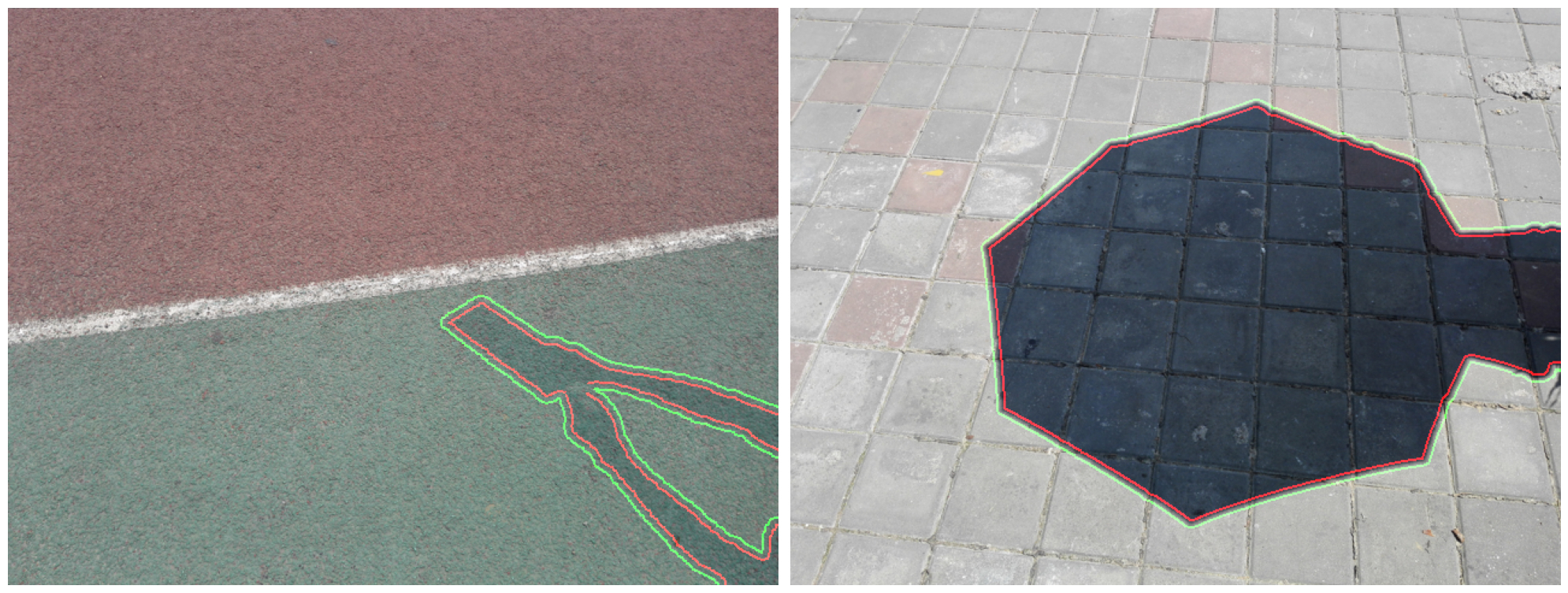}
    \caption{Automatic annotation}
  \end{subfigure}
  \begin{subfigure}{0.48\linewidth}
    \includegraphics[width=\linewidth]{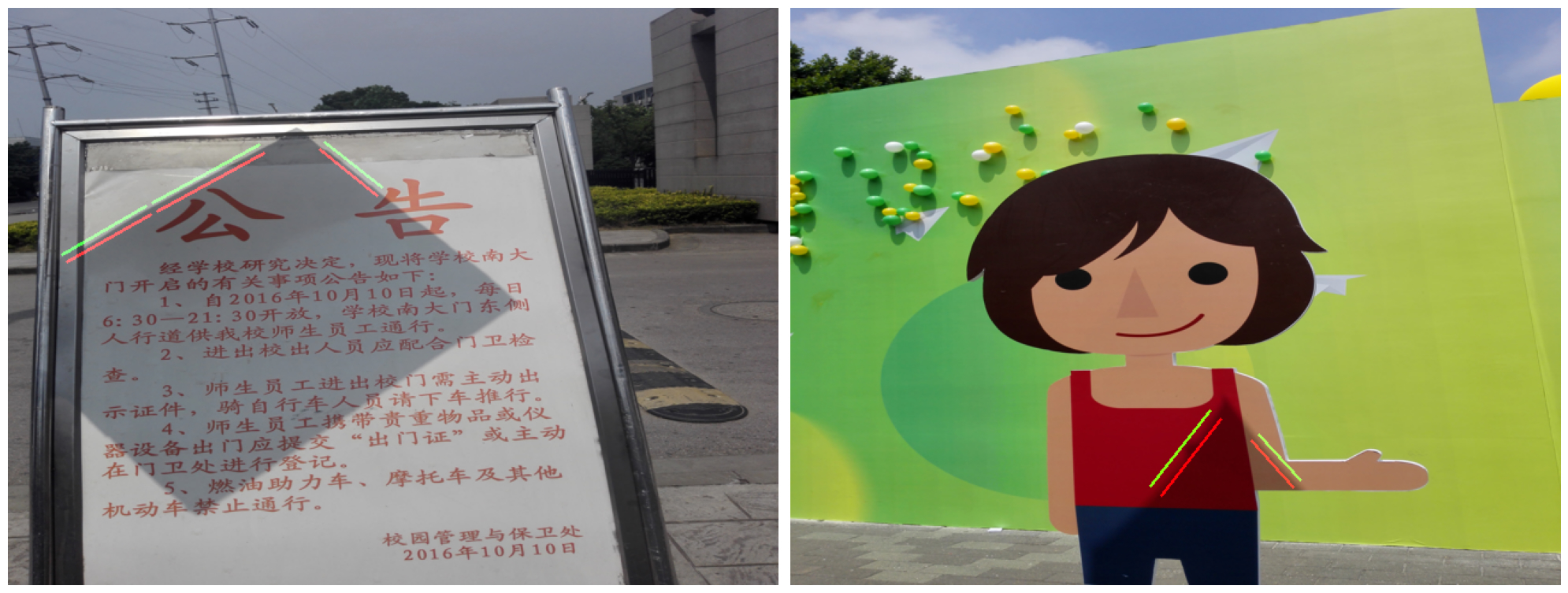}
    \caption{Manual annotation}
  \end{subfigure}
  \caption{Examples of shadow edge pixel annotation in the ISTD+ \cite{Le_2019_ICCV}. (a) shows automatic annotation on 42 scenes using eroded and dilated shadow masks, (b) shows manual annotation of material-consistent pixels on 4 scenes where shadow edges coincide with object boundary.}
  \label{fig: cddanno}
\end{figure}

To provide CDD evaluation on the ISTD+ \cite{Le_2019_ICCV} dataset, we annotate the MC-edges in each image in a semi-automatic manner. Among the 46 unique scenes in the ISTD+ test set, we observe that 42 do not exhibit partial shadow edges coinciding with object boundaries. For these scenes, we simply erode and dilate the original shadow mask and then perform subtraction to get the pixels near the shadow edge. In the remaining four scenes, we manually annotate the MC-edges (as depicted in \cref{fig: cddanno}).

\section{Quantitative results on ISTD+ test set}

Recent improvements in shadow removal on ISTD+ \cite{Le_2019_ICCV} have reached saturation. The test set comprises multiple images from scenes similar to those in the training set, with shadows cast by objects outside the captured scene. SOTA methods effectively learn the mapping between shadow and shadow-free pairs, already yielding satisfactory results.
In \cref{table: istdcdd}, we compare our method with SOTA methods. Our method achieves performance comparable to the SOTA. Although our method improves shadow removal on several challenging cases in ISTD+, the overall performance does not significantly surpass the well-trained existing methods on simple shadow images. This further supports the need to extend shadow removal techniques to general shadow images in real-world scenarios.


\begin{table}[!t]
\centering
\caption{Comparison with SOTA models. We compare the performance of pre-trained models and models using our adaptation method on the ISTD+ \cite{Le_2019_ICCV} dataset. MAE and CDD are reported, note that CDD is reported in $1000\times$ the original value.}
\label{table: istdcdd}
\resizebox{\columnwidth}{!}{%
\begin{tabular}{@{}lccccc@{}}
\toprule
\multicolumn{1}{c}{\multirow{3}{*}{Methods}} & \multicolumn{5}{c}{ISTD+}                                                \\ \cmidrule(l){2-6} 
\multicolumn{1}{c}{}                         & \multicolumn{3}{c}{MAE}                    & \multicolumn{2}{c}{CDD}     \\ \cmidrule(l){2-6} 
\multicolumn{1}{c}{}                         & Shadow       & NonShadow    & All          & Mean         & Var          \\ \midrule
Input                                        & 40.2         & 2.6          & 8.5          & 148.5        & 90.0         \\ \midrule
Inpaint4Shadow\cite{li2023leveraging}        & 5.9          & 2.9          & 3.4          & 2.1          & 4.3          \\ \midrule
ShadowDiffusion\cite{guo2023shadowdiffusion} & \textbf{4.9} & 2.3          & \textbf{2.7} &   /          &  /           \\ \midrule
SP+M-Net\cite{Le_2019_ICCV}                  & 7.3          & 2.5          & 3.3          & 3.2          & 3.8          \\
SP+M-Net+Ours                                & 6.1          & 2.5          & 3.1          & 1.6          & 2.6          \\ \midrule
ShadowFormer\cite{guo2023shadowformer}       & 5.3          & \textbf{2.2} & \textbf{2.7} & 1.5          & 2.6          \\
ShadowFormer+Ours                            & 5.0          & \textbf{2.2} & \textbf{2.7} & \textbf{1.0} & \textbf{1.6} \\ \bottomrule
\end{tabular}%
}
\end{table}

\section{More Results on The Cross-Dataset Testing}
\subsection{Effect of $L_{nonshadow}$}
$L_{nonshadow}$ is specifically designed to mitigate the error caused by prior models pre-trained on data pairs with light intensity inconsistencies. In \cref{tab: nonshadow}, we can see that our proposed method outperforms the pre-trained models, and incorporating $L_{nonshadow}$ further improves the performance by correcting the non-shadow region.

\begin{table}[!ht]
\centering
\caption{Quantitative results on cross dataset testing comparing the effect of $L_{nonshadow}$. ISTD pre-trained ShadowFormer and SRD pre-trained ShadowFormer are tested on the ISTD+ test set.}
\label{tab: nonshadow}
\resizebox{\columnwidth}{!}{%
\begin{tabular}{@{}cclccccc@{}}
\toprule
\multirow{2}{*}{Trained On} & \multirow{2}{*}{Tested On} & \multirow{2}{*}{Methods} & \multicolumn{3}{c}{MAE} & \multicolumn{2}{c}{CDD} \\ \cmidrule(l){4-8} 
                      &                        &                & S    & NS  & A   & Mean & Var  \\ \midrule
\multirow{3}{*}{SRD}  & \multirow{3}{*}{ISTD+} & prior model          & 13.7 & 3.4 & 5.1 & 55.0 & 43.3 \\ \cmidrule(l){3-8} 
                      &                        & w.o. $L_{nonshadow}$ & 6.4  & 2.9 & 3.5 & 15.8 & 13.6 \\ \cmidrule(l){3-8} 
                      &                        & w. $L_{nonshadow}$ & 6.2  & 2.4 & 3.0 & 8.0  & 9.4  \\ \midrule
\multirow{3}{*}{ISTD} & \multirow{3}{*}{ISTD+} & prior model          & 10.6 & 6.3 & 7.0 & 11.8 & 17.7 \\ \cmidrule(l){3-8} 
                      &                        & w.o. $L_{nonshadow}$ & 6.2  & 6.5 & 6.3 & 9.8  & 15.2 \\ \cmidrule(l){3-8} 
                      &                        & w. $L_{nonshadow}$ & 6.3  & 2.7 & 3.4 & 1.0  & 3.1  \\ \bottomrule
\end{tabular}%
}
\end{table}

\subsection{Qualitative results on SRD pre-trained model}
\begin{figure}[!t]
  \centering
  \begin{subfigure}{0.24\linewidth}
    \includegraphics[width=\linewidth]{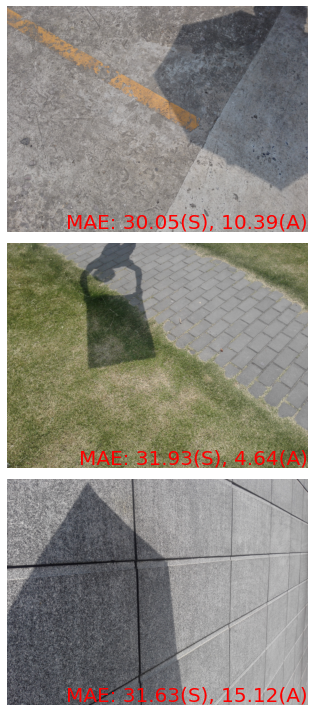}
    \caption{Input}
  \end{subfigure}
  \begin{subfigure}{0.24\linewidth}
    \includegraphics[width=\linewidth]{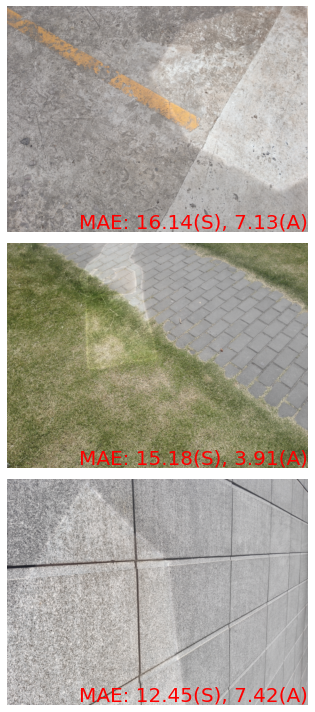}
    \caption{\cite{guo2023shadowformer}}
  \end{subfigure}
  \begin{subfigure}{0.24\linewidth}
    \includegraphics[width=\linewidth]{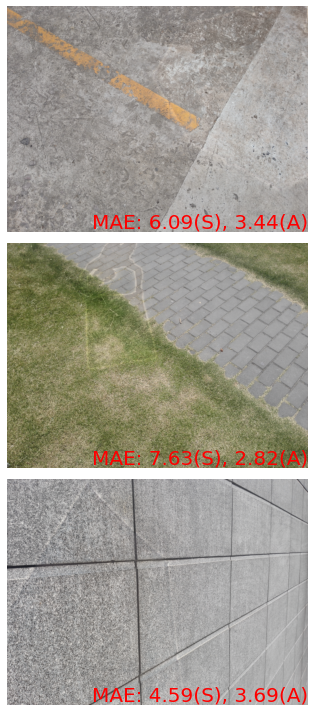}
    \caption{\cite{guo2023shadowformer}+Ours}
  \end{subfigure}
  \begin{subfigure}{0.24\linewidth}
    \includegraphics[width=\linewidth]{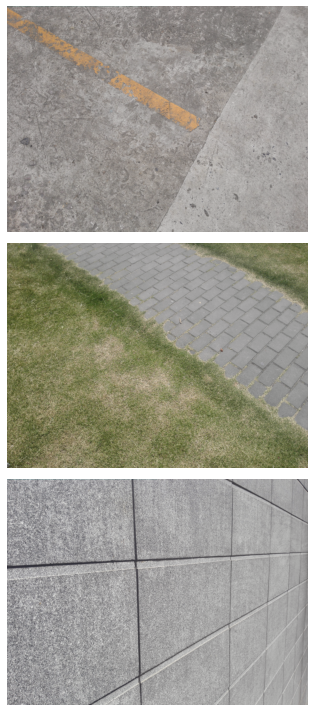}
    \caption{GT}
  \end{subfigure}
  \caption{Qualitative comparison on cross dataset testing. We use the SRD \cite{qu2017deshadownet} pre-trained ShadowFormer \cite{guo2023shadowformer} and test it on the ISTD+ test set. (a) shows input image, (b) shows ShadowFormer result, (c) presents the results with refinement, and (d) presents the ground truth. We also report shadow region (S) and overall (A) MAE results.}
  \label{fig: srd}
\end{figure}

Here we show additional qualitative results of the cross-dataset testing where we use a ShadowFormer \cite{guo2023shadowformer} pre-trained on the SRD \cite{qu2017deshadownet} to test on ISTD+ \cite{Le_2019_ICCV} images with and without our refinement method. As depicted in \cref{fig: srd}, the pre-trained model does not perform well on the out-of-distribution shadow images, while our refinement method significantly improves the performance.

\section{More Qualitative Results}
\label{qual}
\begin{figure}[!ht]
  \centering
  \includegraphics[width=\linewidth]{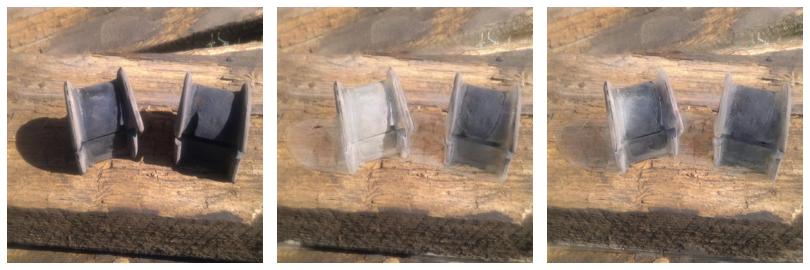}\\
  \includegraphics[width=\linewidth]{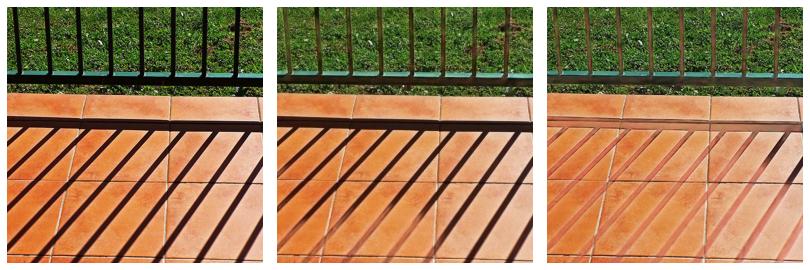}
  \makebox[0.32\linewidth]{\small{Input}}
  \makebox[0.32\linewidth]{\small{\cite{guo2023shadowdiffusion}}}
  \makebox[0.32\linewidth]{\small{\cite{guo2023shadowdiffusion}+Ours}}
  \caption{Qualitative results on our proposed dataset. (a) shows input image, (b) shows ShadowDiffusion \cite{guo2023shadowdiffusion} result, and (c) presents the results after refinement.}
  \label{fig: diffusion}
\end{figure}

\begin{figure*}[!ht]
\centering
\includegraphics[width=\linewidth]{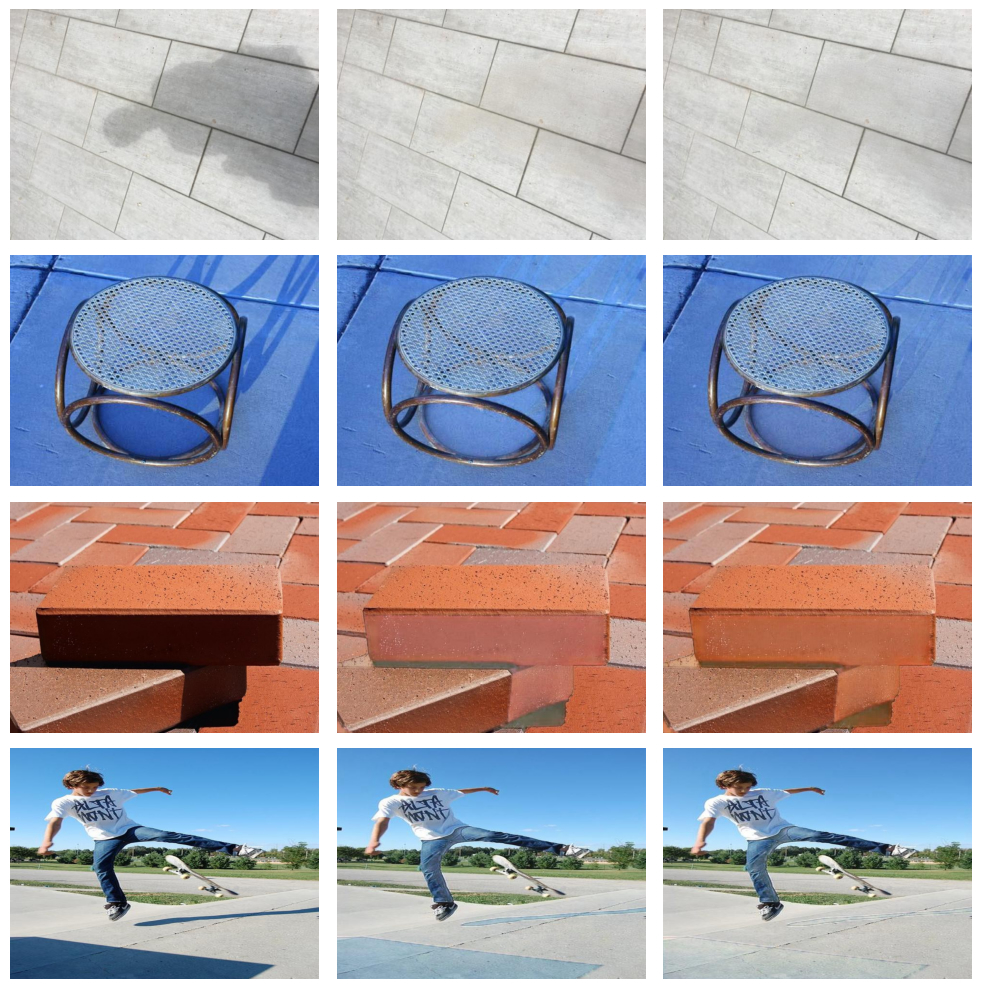}
\makebox[0.32\linewidth]{\small{(a) Input}}
\makebox[0.32\linewidth]{\small{(b) ShadowFormer}}
\makebox[0.32\linewidth]{\small{(c) ShadowFormer+Ours}}
\caption{Qualitative results on soft shadows and self-cast shadows. The top two rows show results on soft shadows which are easier cases for the prior model \cite{guo2023shadowformer}, and the bottom two rows show results on self-cast shadows which confuses the prior model. However, our method also cannot fully address the attached shadows.}
\label{fig: type}
\vspace{-3mm}
\end{figure*}

\begin{figure*}
  \centering
  \begin{subfigure}{0.3\linewidth}
    \includegraphics[width=\linewidth]{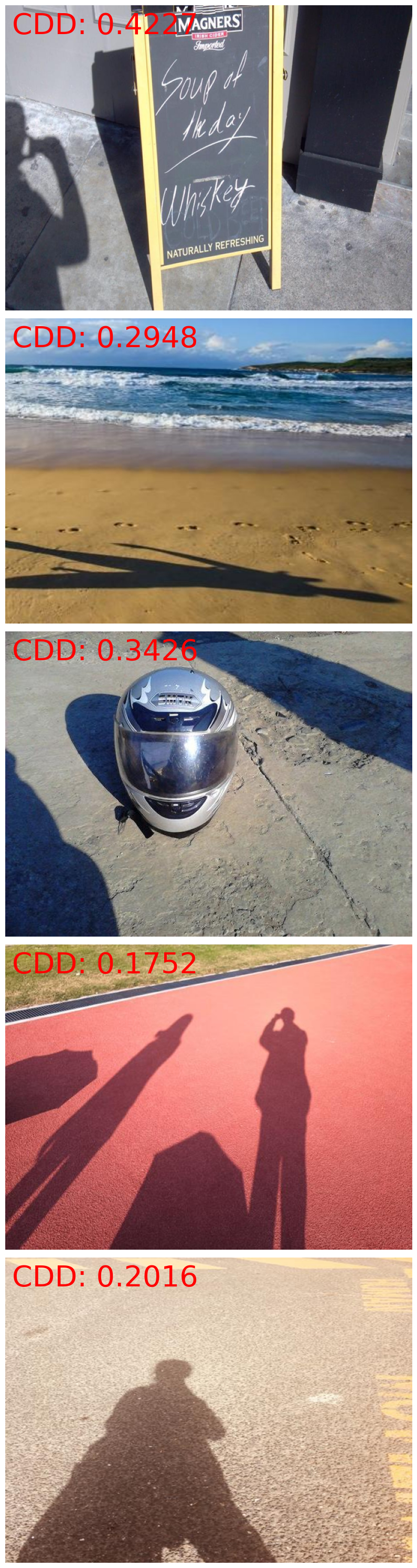}
    \caption{Input}
  \end{subfigure}
  \begin{subfigure}{0.3\linewidth}
    \includegraphics[width=\linewidth]{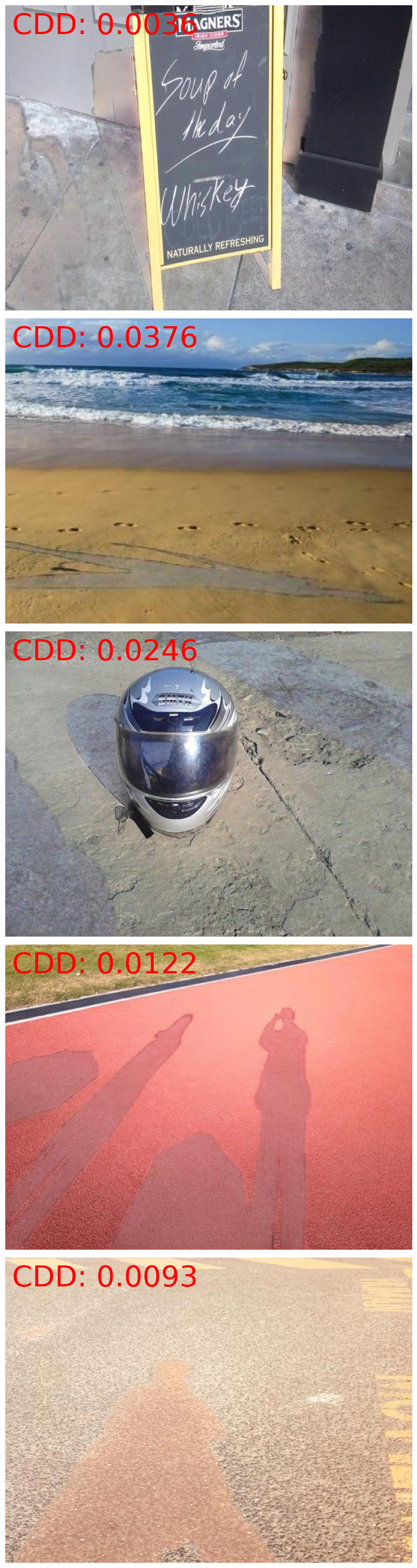}
    \caption{ShadowFormer}
  \end{subfigure}
  \begin{subfigure}{0.3\linewidth}
    \includegraphics[width=\linewidth]{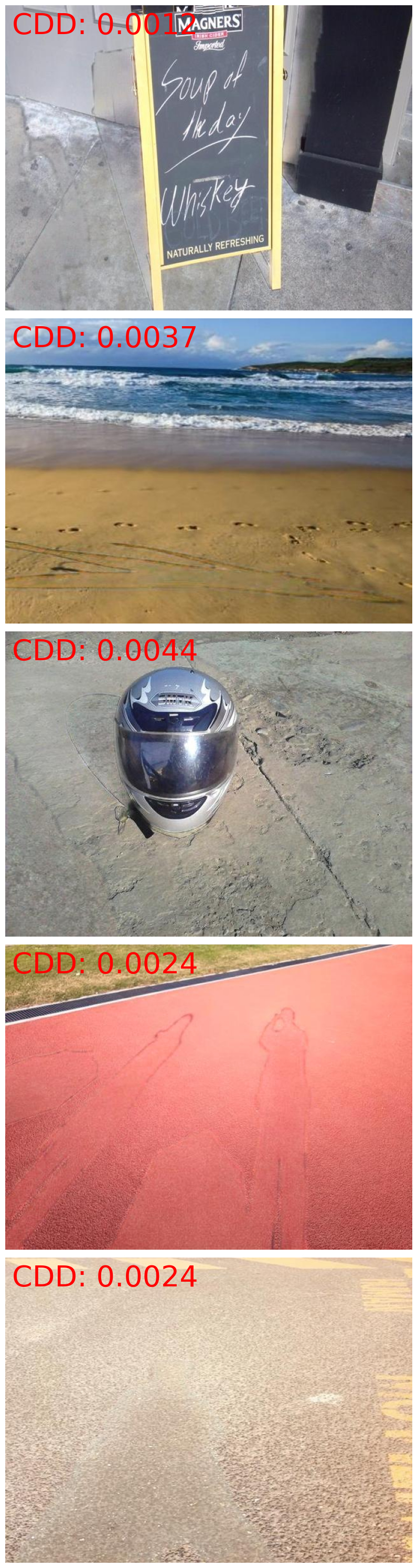}
    \caption{ShadowFormer+Ours}
  \end{subfigure}
  \caption{Qualitative results on our proposed dataset. (a) shows input image, (b) shows ShadowFormer \cite{guo2023shadowformer} result, and (c) presents the results after refinement.}
  \label{fig: supproposedsf}
\end{figure*}

\begin{figure*}
  \centering
  \begin{subfigure}{0.3\linewidth}
    \includegraphics[width=\linewidth]{pics_supp/supproposed1.jpg}
    \caption{Input}
  \end{subfigure}
  \begin{subfigure}{0.3\linewidth}
    \includegraphics[width=\linewidth]{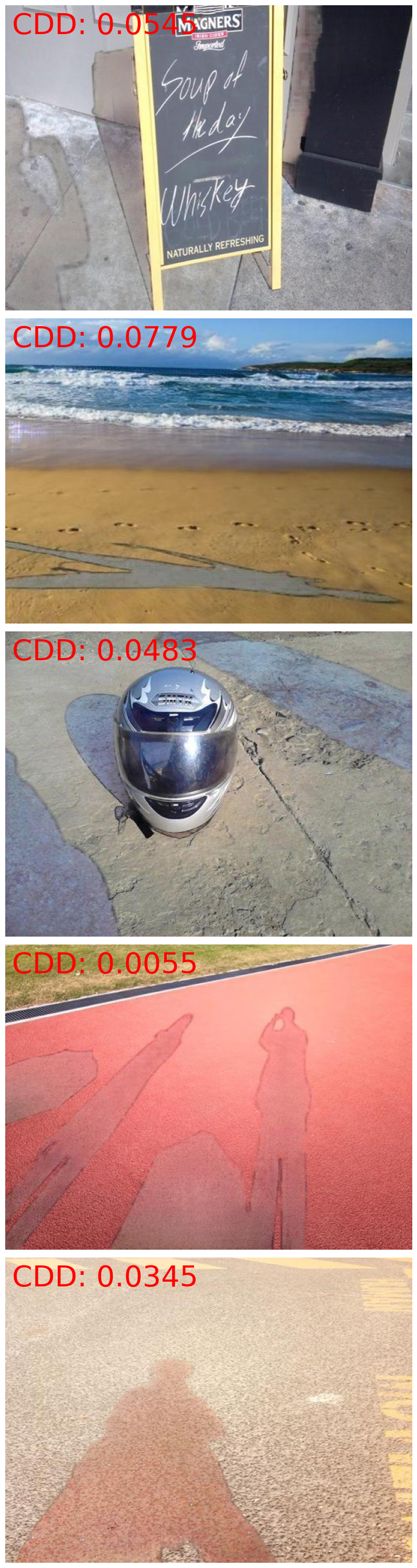}
    \caption{SP+M-Net}
  \end{subfigure}
  \begin{subfigure}{0.3\linewidth}
    \includegraphics[width=\linewidth]{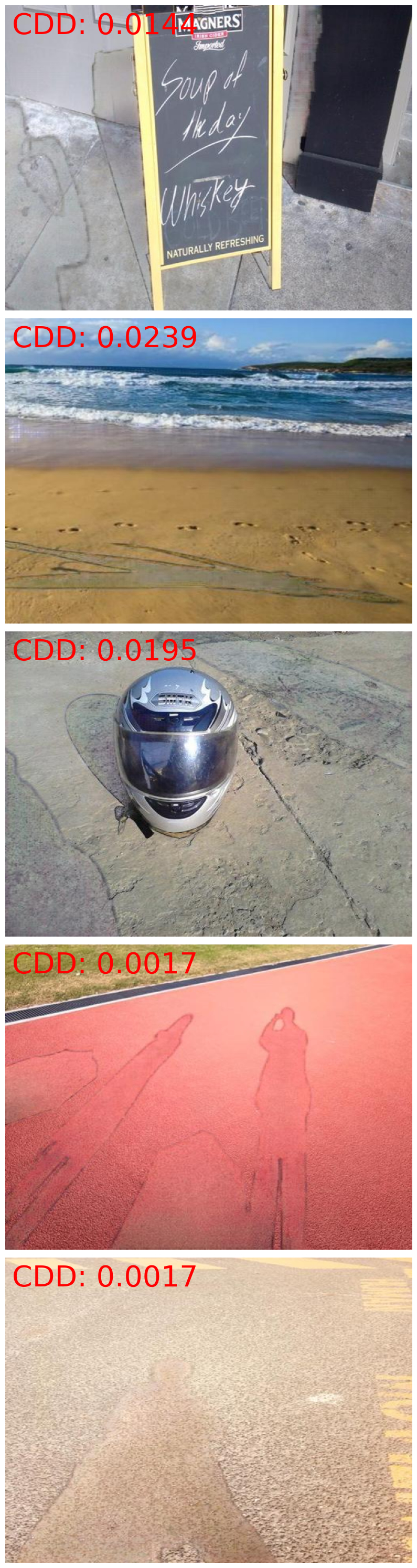}
    \caption{SP+M-Net+Ours}
  \end{subfigure}
  \caption{Qualitative results on our proposed dataset. (a) shows input image, (b) shows SP+M-Net \cite{Le_2019_ICCV} result, and (c) presents the results after refinement.}
  \label{fig: supproposedspm}
\end{figure*}

\begin{figure*}
  \centering
  \begin{subfigure}{0.24\linewidth}
    \includegraphics[width=\linewidth]{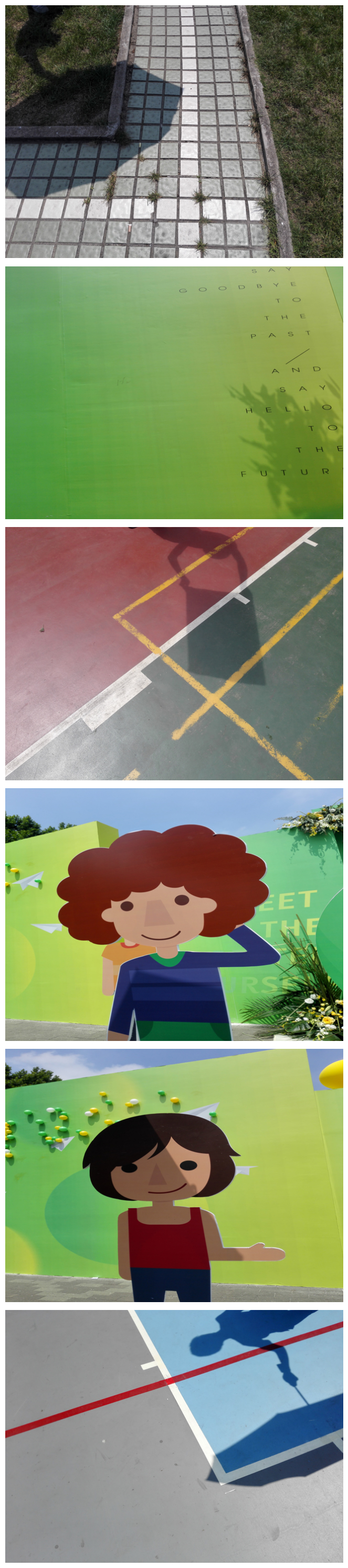}
    \caption{Input}
  \end{subfigure}
  \begin{subfigure}{0.24\linewidth}
    \includegraphics[width=\linewidth]{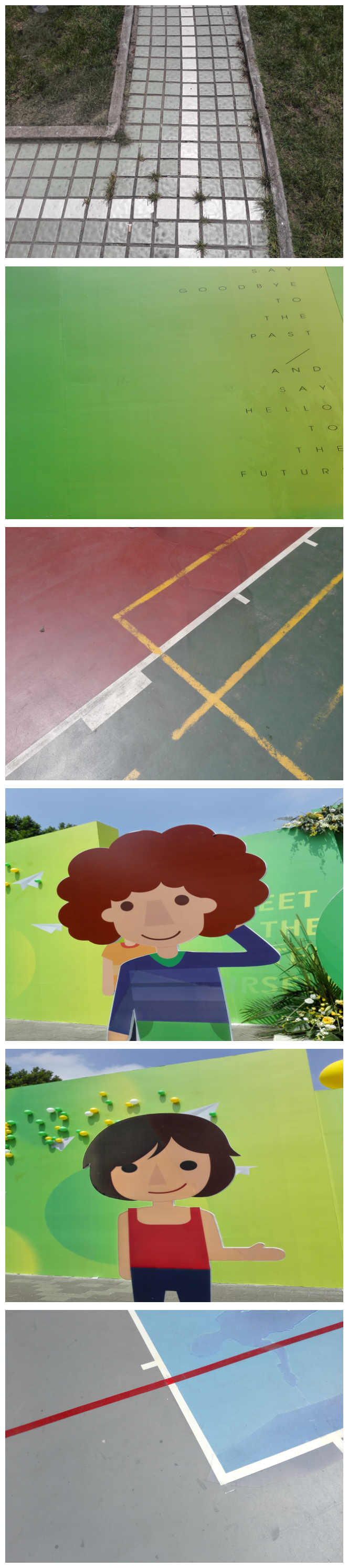}
    \caption{\cite{guo2023shadowformer}}
  \end{subfigure}
  \begin{subfigure}{0.24\linewidth}
    \includegraphics[width=\linewidth]{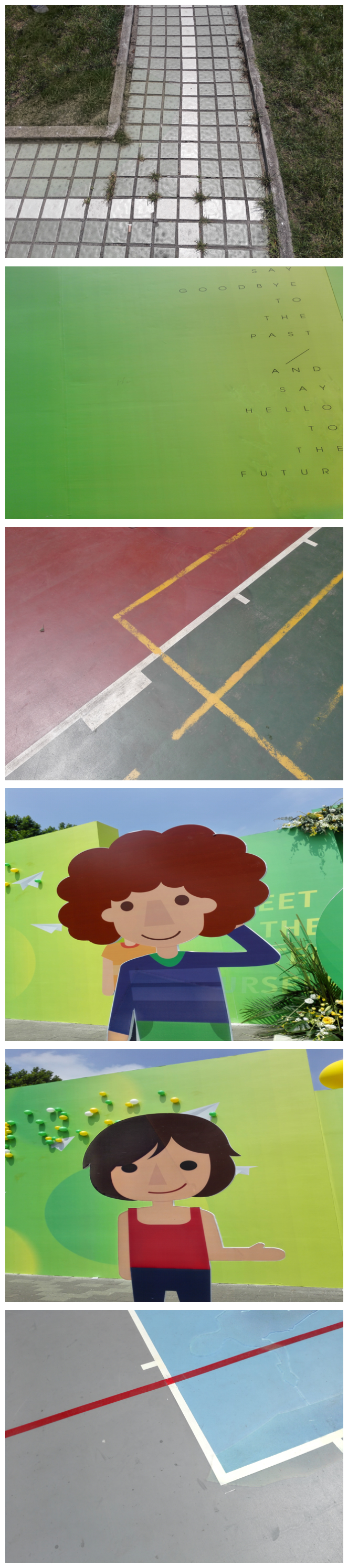}
    \caption{\cite{guo2023shadowformer}+Ours}
  \end{subfigure}
  \begin{subfigure}{0.24\linewidth}
    \includegraphics[width=\linewidth]{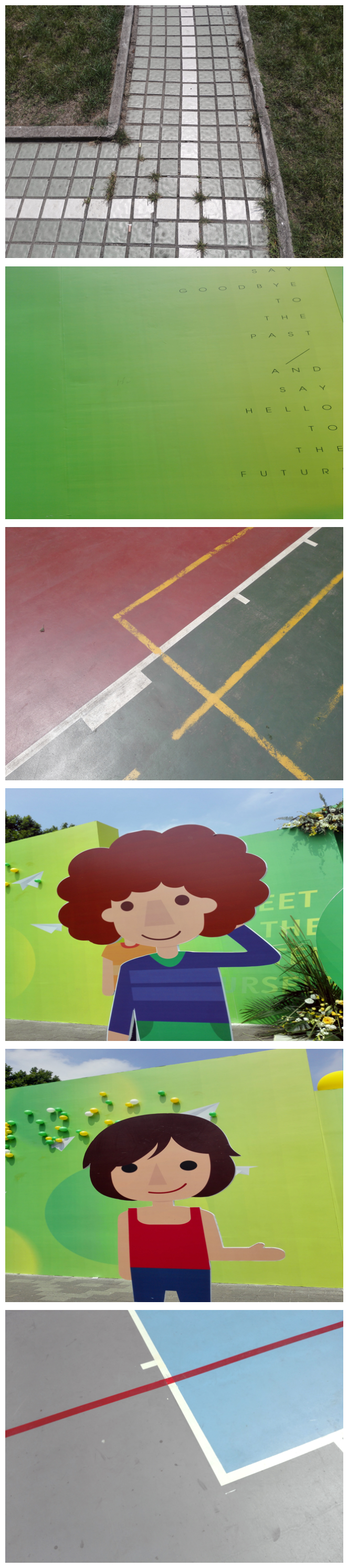}
    \caption{GT}
  \end{subfigure}
  \caption{More qualitative results on ISTD+ \cite{Le_2019_ICCV}. (a) shows input image, (b) shows ShadowFormer \cite{guo2023shadowformer} result, (c) presents the results using our refinement method, and (d) shows the ground truth.}
  \label{fig: supistd}
\end{figure*}

Qualitative results on refining a pre-trained diffusion-based model are provided in \cref{fig: diffusion}.
We present the qualitative results on soft shadows and self-cast shadows in \cref{fig: type}. 
We also show more qualitative results on both our proposed dataset and the ISTD+ \cite{Le_2019_ICCV} in \cref{fig: supproposedsf}, \cref{fig: supproposedspm}, and \cref{fig: supistd}.

\end{document}